\documentclass{article}

    \PassOptionsToPackage{numbers, compress}{natbib}

     \usepackage[preprint]{tackling_climate_workshop_style}

\usepackage[utf8]{inputenc} 
\usepackage[T1]{fontenc}    
\usepackage{hyperref}       
\usepackage{url}            
\usepackage{booktabs}       
\usepackage{amsfonts}       
\usepackage{nicefrac}       
\usepackage{microtype}      
\usepackage{amsmath, amssymb, amsthm}
\usepackage{graphicx}
\usepackage{caption}
\usepackage{subcaption}
\usepackage{algorithm}
\usepackage{algpseudocode}

\theoremstyle{plain}
\newtheorem{example}{Example}

\renewcommand{\vec}[1]{\boldsymbol{#1}}
\newcommand{\mat}[1]{\mathbf{#1}}
\def\diff{\mathrm{d}}

\title{Short-term Prediction and Filtering of Solar Power Using State-Space Gaussian Processes}

\author{%
  Sean Nassimiha\thanks{Also with Balyasny Asset Management} \\
  Department of Computer Science\\
  University College London \\
  \texttt{sean.nassimiha.18@ucl.ac.uk} \\
   \AND
   Peter Dudfield \\
   Open Climate Fix \\
   \texttt{peter@openclimatefix.org} \\
   \And
   Jack Kelly \\
   Open Climate Fix \\
   \texttt{jack@openclimatefix.org} \\
   \And
   Marc Peter Deisenroth \\
   UCL Centre for Artificial Intelligence \\
   University College London \\
   \texttt{m.deisenroth@ucl.ac.uk} \\
   \And
   So Takao \\
   UCL Centre for Artificial Intelligence \\
   University College London \\
   \texttt{so.takao@ucl.ac.uk} \\
}

\begin{document}

\maketitle

\begin{abstract}
Short-term forecasting of solar photovoltaic energy (PV) production is important for powerplant management. Ideally these forecasts are equipped with error bars, so that downstream decisions can account for uncertainty. To produce predictions with error bars in this setting, we consider Gaussian processes (GPs) for modelling and predicting solar photovoltaic energy production in the UK. A standard application of GP regression on the PV timeseries data is infeasible due to the large data size and non-Gaussianity of PV readings. However, this is made possible by leveraging recent advances in scalable GP inference, in particular, by using the state-space form of GPs, combined with modern variational inference techniques. The resulting model is not only scalable to large datasets but can also handle continuous data streams via Kalman filtering.
\end{abstract}

\section{Introduction}

On Tuesday, 19 July 2022, Britain paid the highest electricity price ever recorded, at a staggering
£9724 per megawatt hour. This (quite horrible) achievement was due to a deep imbalance in
the supply and demand of the national power grid, which was under severe stress caused by the
hottest UK day on record, and by the recent surge in natural gas prices~\cite{bbc_article}.
It is apparent that our continual reliance on natural gas had enabled this perfect storm; it can be argued that our energy system may have been more resilient had there been a larger proportion of renewables in the mix.

\begin{figure}[h]
    \centering
    \begin{subfigure}[b]{0.22\textwidth}
         \centering
         \includegraphics[width=\textwidth]{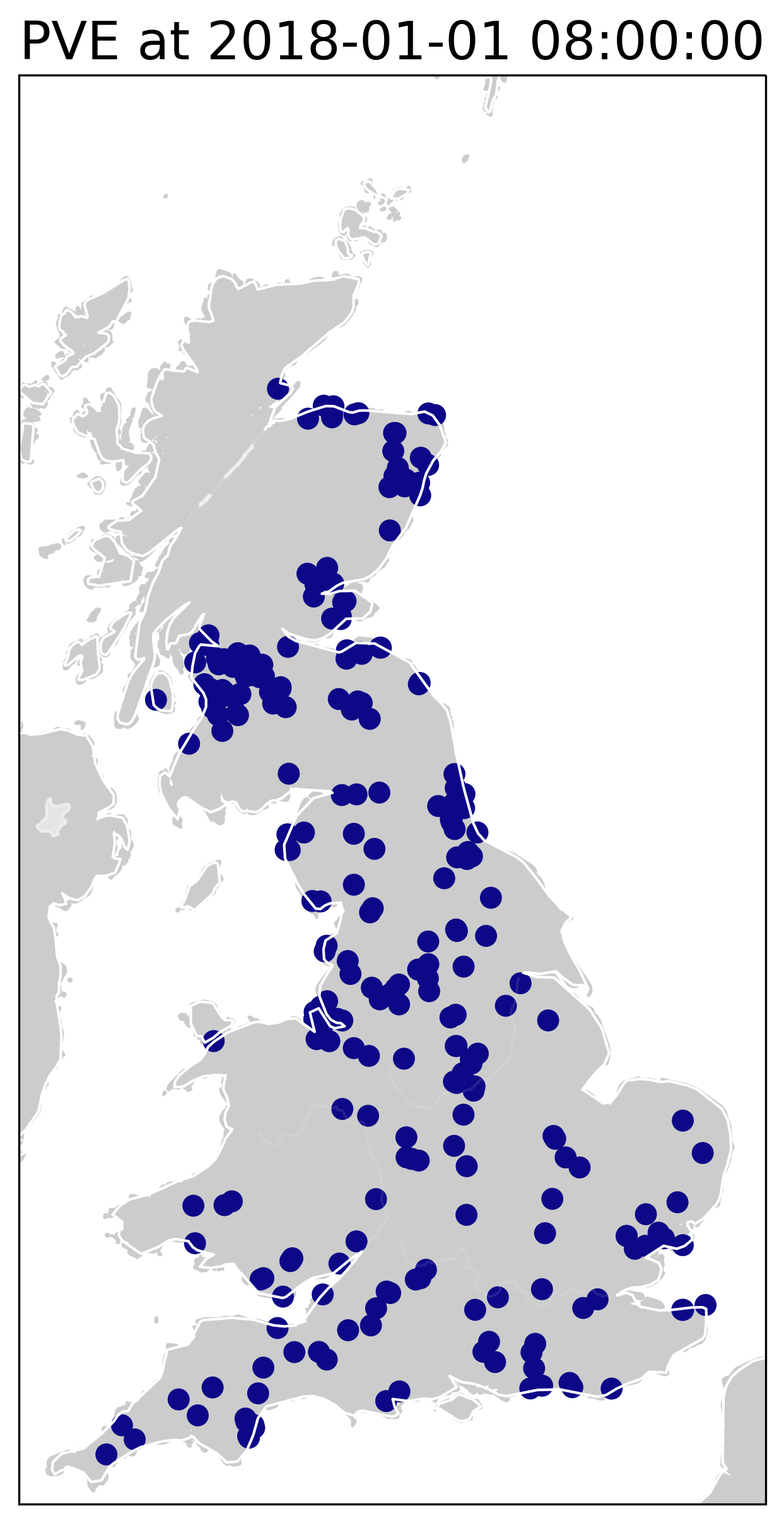}
     \end{subfigure}
     \begin{subfigure}[b]{0.22\textwidth}
         \centering
         \includegraphics[width=\textwidth]{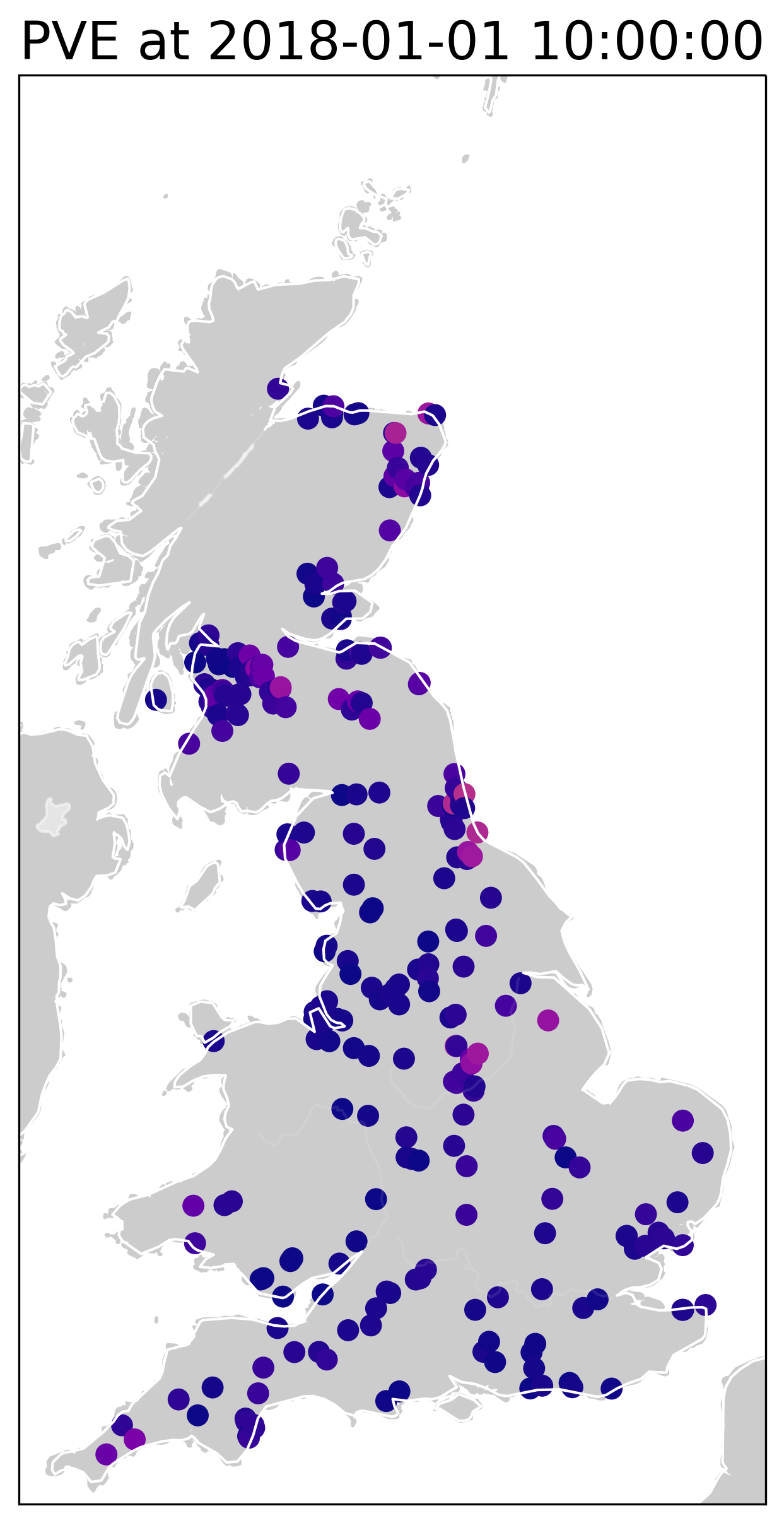}
     \end{subfigure}
     \begin{subfigure}[b]{0.22\textwidth}
         \centering
         \includegraphics[width=\textwidth]{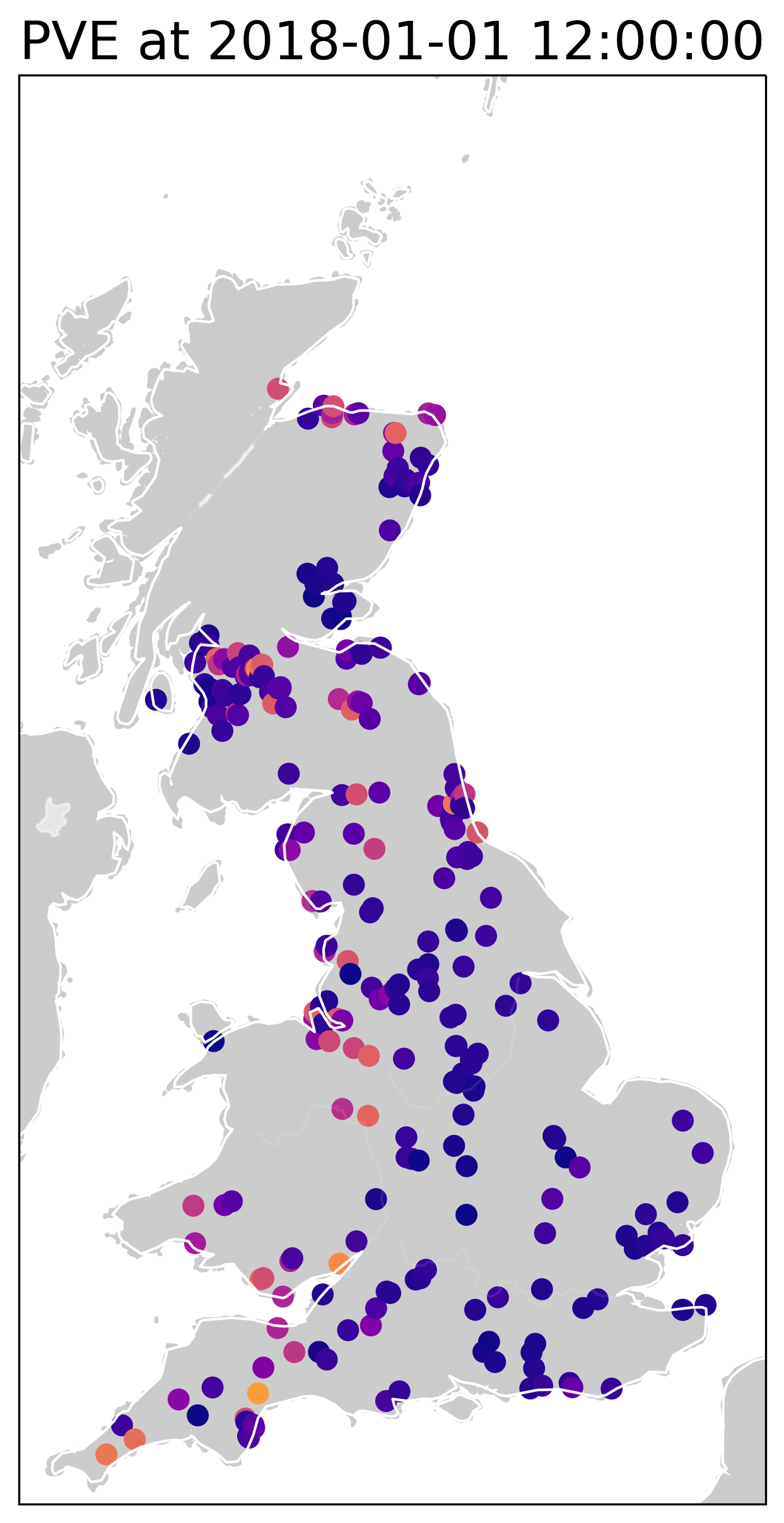}
     \end{subfigure}
     \begin{subfigure}[b]{0.22\textwidth}
         \centering
         \includegraphics[width=\textwidth]{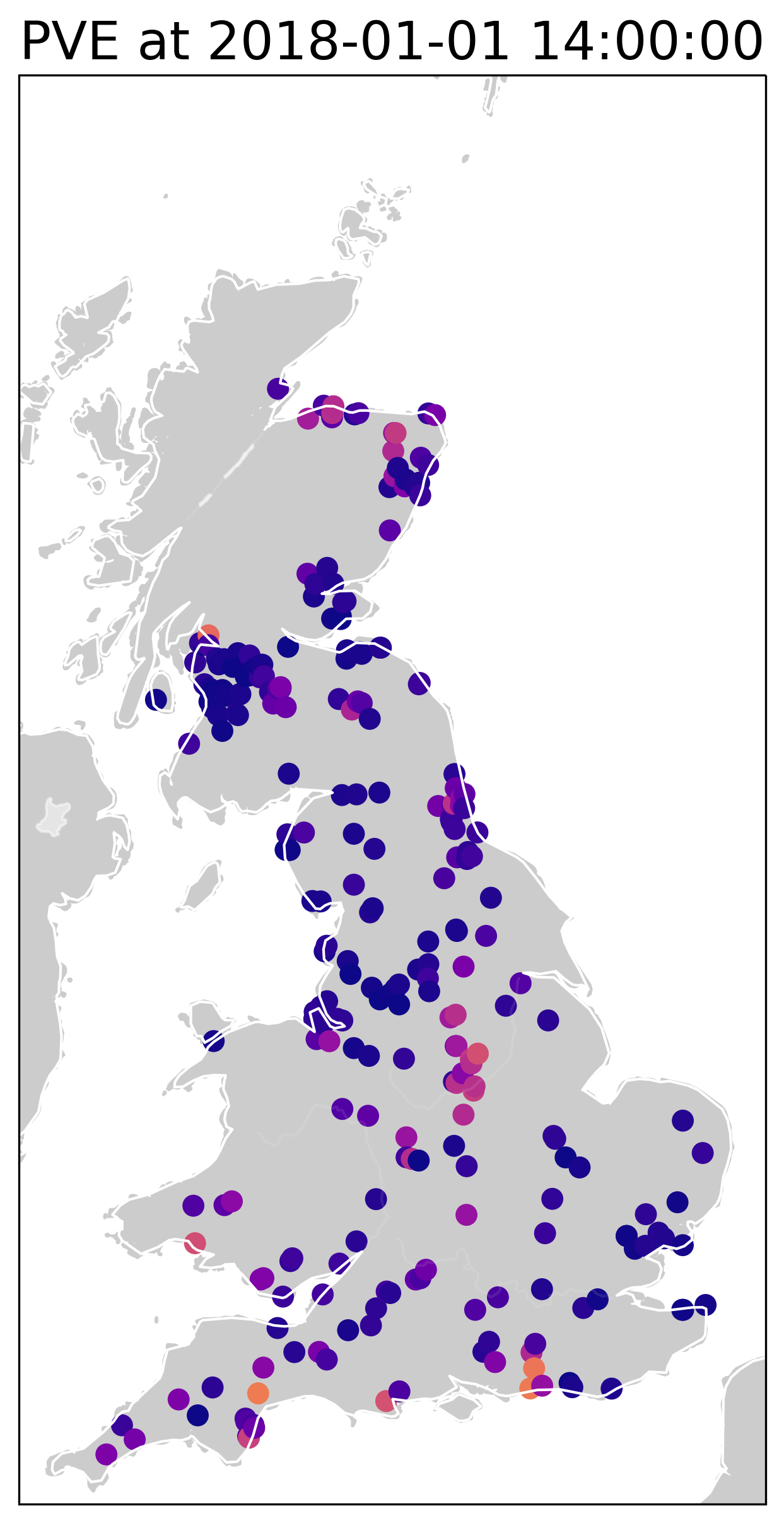}
     \end{subfigure}
    \caption{PV output at stations distributed across the UK at different times.}
    \label{fig:pv-stations}
\end{figure}

However, managing supply and demand through renewables is not easy. Solar power, among most other renewables, is intermittent, making it necessary to back it up with spinning reserves to provide energy when the sun is not shining. These spinning reserves are, unfortunately, expensive and emit large amounts of CO2~\cite{ocf_nowcasting}.
Thus, by increasing the accuracy of forecasts for renewable electricity supply, one can hope to reduce the amount of spinning reserves and in turn, decrease emissions and costs associated
with the use of clean energy sources.

Naturally, there has been a surge of recent works applying machine learning to the task of photovoltaic energy (PV) nowcasting. For example in \cite{zhang2018deep}, a comparative study of several deep learning models for the task of one-minute ahead PV value prediction is conducted. Similarly, in \cite{luo2021deep}, a physics-constrained LSTM architecture is considered for one-hour ahead PV prediction  
(see also \cite{garud2021review, su2019machine} for comprehensive reviews on the application of ML models for PV nowcasting). While these deep learning models outperform simple baselines, such as persistence, they do not quantify uncertainty, which we believe is necessary for high-stakes applications such as this.

In this paper, we propose the use of Gaussian processes (GPs) for PV nowcasting as alternative to deep-learning based approaches. These are nonparametric Bayesian regression method that can produce predictions equipped with uncertainty estimates~\cite{williams2006gaussian}, which could be useful in our setting.




\section{Data}\label{sec:data}
The dataset that we use in this work was provided to us by Open Climate Fix~\footnote{\url{https://huggingface.co/datasets/openclimatefix/uk_pv}}, consisting of solar PV energy (PVE) readings at 5 minute intervals for 1311 PV stations scattered across Great Britain, between January 2018 and November 2021 (see Figure~\ref{fig:pv-stations}).

\begin{figure}[h]
    \centering
    \begin{subfigure}[b]{0.48\textwidth}
         \centering
         \includegraphics[width=\textwidth]{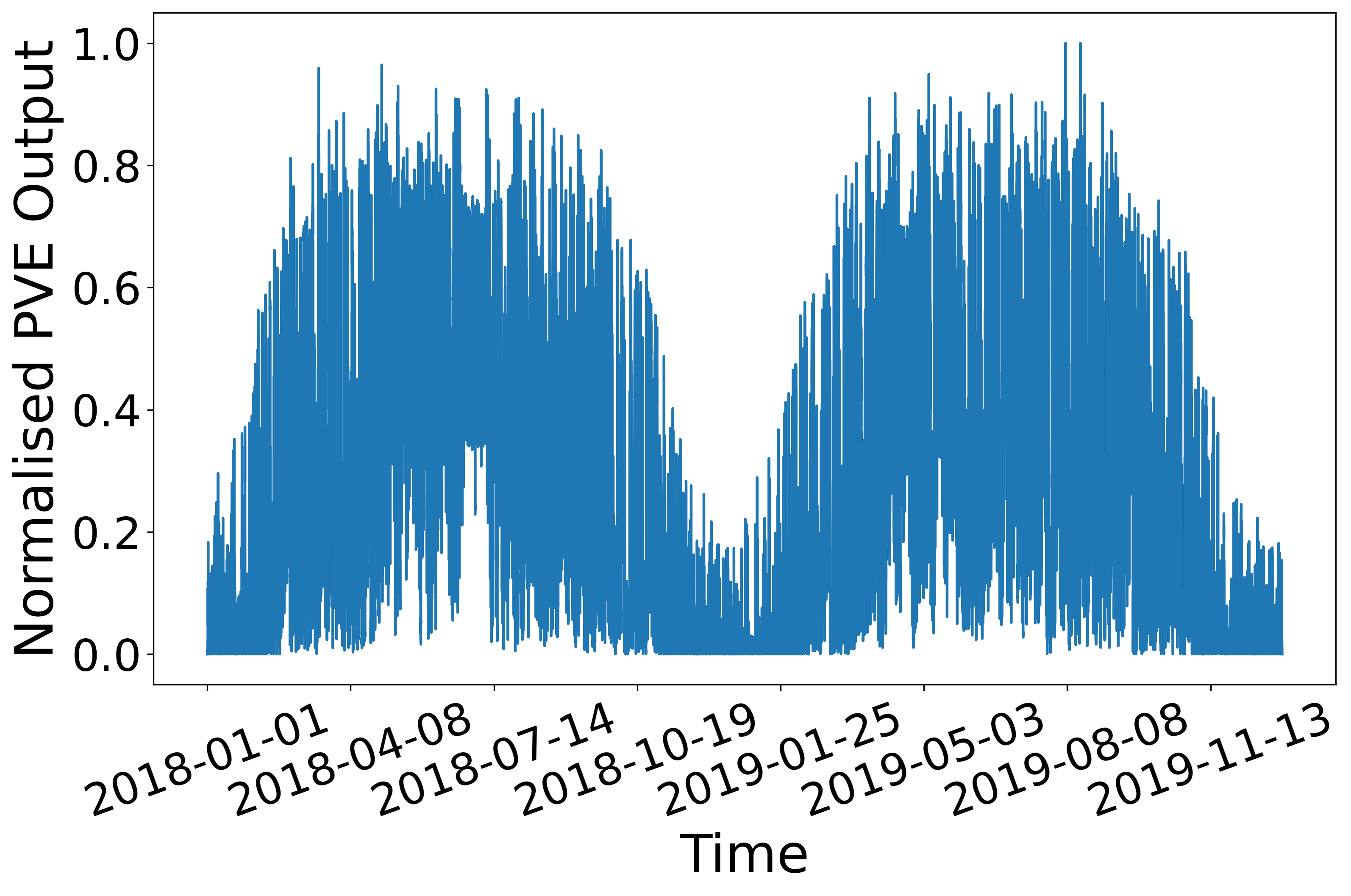}
     \end{subfigure}
     \hfill
     \begin{subfigure}[b]{0.48\textwidth}
         \centering
         \includegraphics[width=\textwidth]{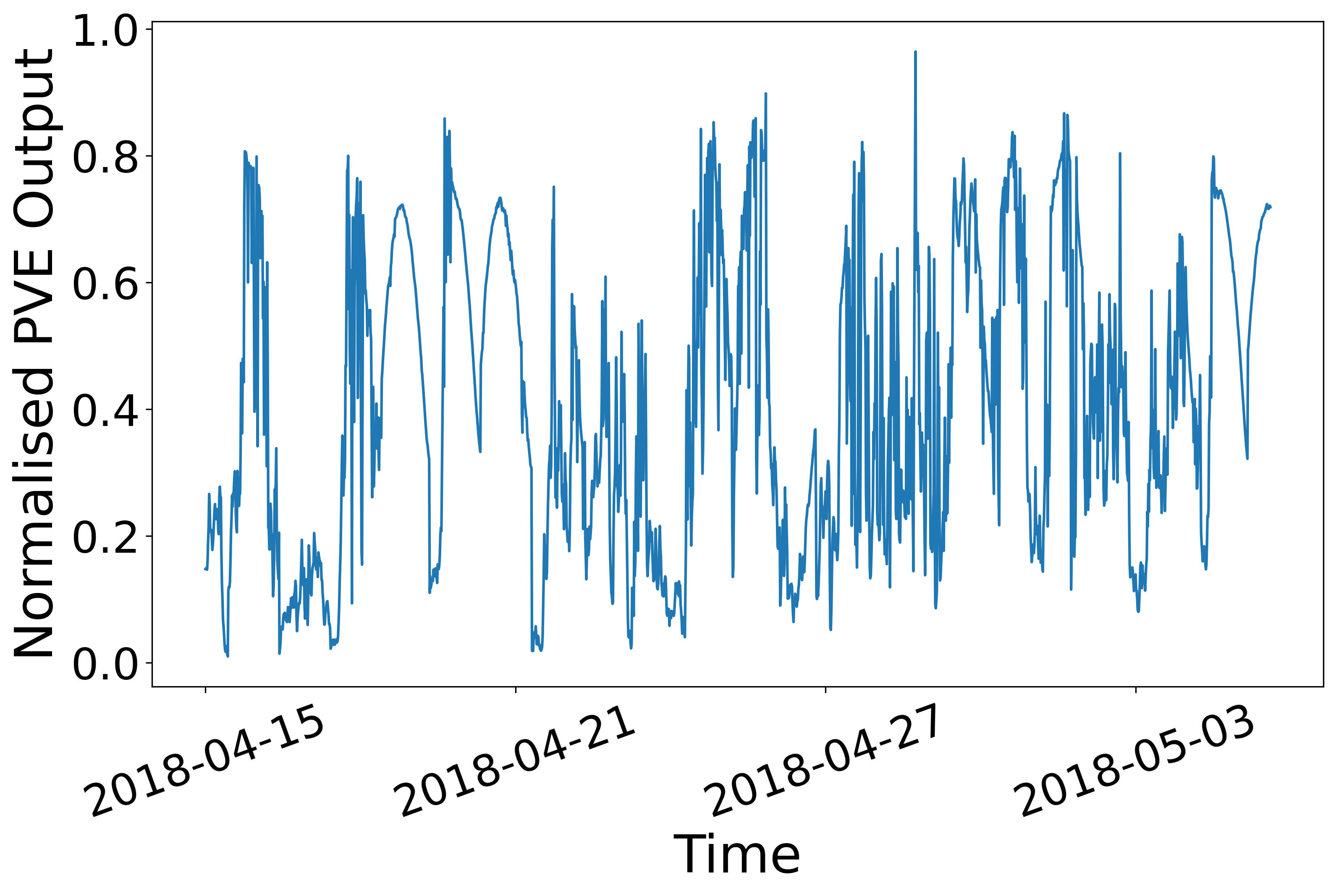}
     \end{subfigure}
    \caption{PV timeseries plot at annual and daily scales. The seasonality at the two scales are apparent.}
    \label{fig:example-timeseries}
\end{figure}

We preprocess the data by first clipping the PV values between $0$\,W and $5 \times 10^7$\,W to remove outliers. This was then re-scaled by dividing each observation by the total capacity of the corresponding PV station, resulting in readings between $[0, 1]$. Data from stations that showed non-zero production overnight or contained missing observations were removed. Finally, the time series were sliced to only include data between 8:00 and 16:00.
The resulting timeseries $(y_i)_{i=1}^N$ at a single station is displayed in Figure~\ref{fig:example-timeseries}, which shows strong annual and daily seasonality.

\section{Methodology} \label{sec:method}
To keep this paper self-contained, we supplement the readers with necessary background on GPs and details regarding our methodology in Appendix~\ref{app:background}.

\subsection{Model} \label{sec:model}
To benchmark our results on the PV nowcasting task, we consider two GPs to model the latent timeseries $f(t)$: the first is a plain Mat\'ern-3/2 GP and in the second, we add a quasi-periodic component, which itself is a product between a periodic kernel and a Matérn-3/2 kernel:
\begin{align}
    k_{\text{model 1}}(t, t') &= k_{\text{mat\'ern-}3/2}(t, t'), \label{eq:model-1} \\
    \begin{split}
    k_{\text{model 2}}(t, t') &= k_{\text{mat\'ern-}3/2}(t, t') + k_{\text{mat\'ern-}3/2}(t, t') \, k_{\text{periodic}}(t, t').
    \end{split}\label{eq:model-2}
\end{align}

Given the latent timeseries $f(\cdot) \sim \mathcal{GP}(0, k(\cdot, \cdot))$, we consider a beta likelihood
\begin{align}\label{eq:likelihood}
    &p(y | f) = \prod_{i=1}^N \mathcal{B}\big(y_i \,|\, \alpha(f(t_i)), \beta(f(t_i))\big), \\
    \text{where} \quad &\alpha(f) = \Phi^{-1}(f) S_B, \quad \beta(f) = S_B - \alpha(f).
\end{align}
Here, $\Phi^{-1} : \mathbb{R} \rightarrow \mathbb{R}$ is the probit transformation and $S_B$ is the scale hyperparameter. We chose this likelihood since the beta distribution $\mathcal{B}$ is defined over the interval $[0, 1]$ and is capable of modelling a wide variety of shapes.
This is suitable in our setting where the preprocessed data $(y_i)_{i=1}^N$ take values in $[0, 1]$ and can have a skew towards $0$ at certain times of the day.

We also consider a variety of commonly used baselines to assess the performance of our GP models \eqref{eq:model-1}--\eqref{eq:model-2} on the nowcasting task. These include the persistence, yesterday, hourly smoothing, exponential smoothing and vector autoregressive models. Details can be found in Appendix \ref{app:baselines}.

\subsection{State-space representation}
One year of PV measurements from a single station contains $\mathcal{O}(10^5)$ data points, which is too large for a straight-forward application of GP regression. To deal with this, we follow the works~\cite{hartikainen2010kalman, sarkka2013spatiotemporal} to reformulate the system \eqref{eq:model-1}+\eqref{eq:likelihood} or \eqref{eq:model-2}+\eqref{eq:likelihood} as a {\em state-space model}
\begin{align}
    \vec{x}_{t+1} &= \mat{A}_t \vec{x}_t + \vec{\varepsilon}_t, \quad \vec{\varepsilon}_t \sim \mathcal{N}(\vec 0, \mat{Q}_t), \label{eq:dynamics-model} \\
    y_t &= \mat{H}_t \vec{x}_t + \eta_t, \quad \eta_t \sim \mathcal{N}(\vec 0, R_t), \label{eq:observation-model}
\end{align}
which enables us to use the Kalman filter (KF) to infer the posterior $p(\vec{x}_{t}|\vec{y}_{1:t})$.
Since our likelihood~\eqref{eq:likelihood} is non-Gaussian, we use conjugate-computation variational inference~\cite{chang2020fast, khan2017conjugate} to approximately represent it in the form~\eqref{eq:observation-model}, which we explain in Appendix~\ref{app:VI}.
The cost of KF is linear in data size as opposed to the cubic cost of vanilla GP regression, making the inference tractable.
Moreover, this has the added advantage of being able to consume new data on-the-fly without having to retrain.

\section{Results} \label{sec:results}
We now demonstrate our results on the solar PV nowscasting task.
We used the {\tt BayesNewton} package\footnote{https://github.com/AaltoML/BayesNewton} to implement the experiments using state-space GP. All experiments were conducted on a MacBook Pro with Apple M1 chip and 16 GB memory.

We assess the performance of our model on a set of 27 PV stations in South-East England by cross-validating on 78 different dates between 2018-2019 (see Appendix~\ref{app:cross-validation} for details on the cross-validation). In each CV fold, we use a period of 100 days for training and the subsequent two hours for prediction. For the metric, we use the mean absolute error and the negative log predictive density (see Appendix~\ref{app:evaluation-metric}), averaged over the 27 systems. We report the average values and variability of the MAE and NLPD across the 78 CV folds in Table~\ref{table:final-results}.

\begin{table}[h]
\centering
\begin{tabular}{ |c|c|c| } 
 \hline
 Model & MAE $\downarrow$ (mean $\pm$ std) & NLPD $\downarrow$ (median $\pm$ m.a.d.) \\ 
 \hline
 Persistence & $0.119 \pm 0.060$ & N/A \\ 
 Yesterday & $0.152 \pm 0.091$ & N/A \\ 
 Hourly smoothing & $0.125 \pm 0.061$ & N/A \\ 
 Simple exponential smoothing & $0.117 \pm 0.058$ & $-11.1 \pm 11.1$ \\ 
 Seasonal exponential smoothing & $0.110 \pm 0.049$ & $-12.2 \pm 10.4$ \\
 Vector autoregression & $0.129 \pm 0.071$ & N/A \\ 
 \hline
 SS-GP model 1 & $0.134 \pm 0.056$ & $-4.93 \pm 16.6$ \\ 
 SS-GP model 2 & $\mathbf{0.109 \pm 0.050}$ & $\mathbf{-12.9 \pm 13.8}$ \\ 
 \hline
\end{tabular}
\caption{Results of the GP models and baselines on the two-hour-ahead PV prediction task. We display the mean $\pm$ standard deviation of the MAE and median $\pm$ median absolute deviation of the NLPD across the 78 CV folds. The median was used for the NLPD due to the presence of outliers.}
\label{table:final-results}
\end{table}

Our cross-validation results suggest that model 2 performs the best overall, both in terms of the MAE and NLPD, beating the seasonal exponential smoothing baseline by a margin. On the other hand, our control model 1 performs poorly, beating only the worst-performing `yesterday' baseline. This indicates the importance of including periodicity to get good predictions.

\begin{figure}[h]
    \centering
    \begin{subfigure}[b]{0.48\textwidth}
         \centering
         \includegraphics[width=\textwidth]{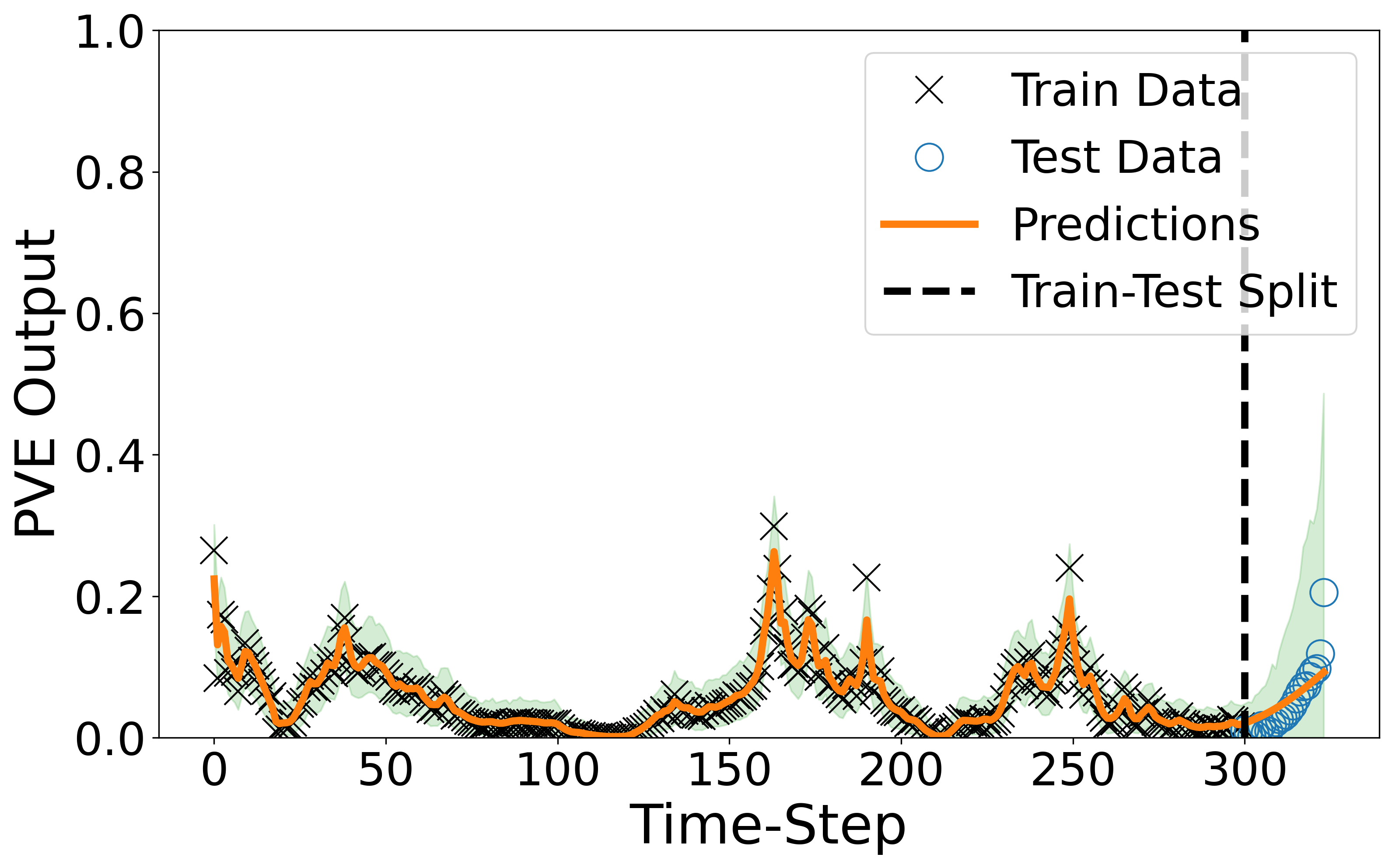}
     \end{subfigure}
     \hfill
     \begin{subfigure}[b]{0.48\textwidth}
         \centering
         \includegraphics[width=\textwidth]{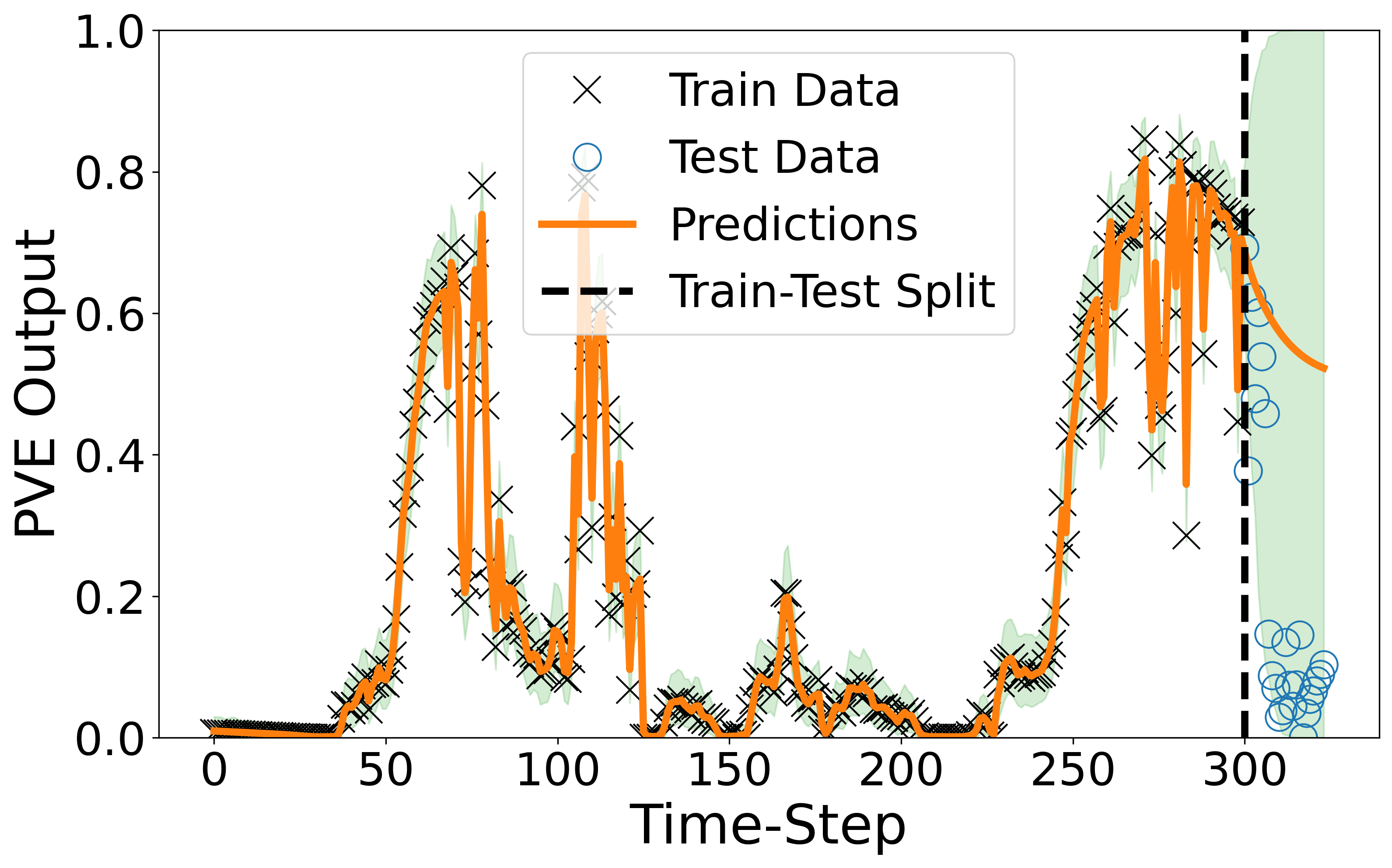}
     \end{subfigure}
    \caption{Example predictions from model 2 equipped with $95\%$ credible intervals on two dates: 2018-01-29 (left) and 2018-02-01 (right) at a given PV station. Predictions are poor when cloud cover is sporadic (right), due to higher PVE volatility. See also Figure~\ref{fig:additional-predictions} for additional plots.}
    \label{fig:example-predictions}
\end{figure}

In Figure~\ref{fig:example-predictions}, we plot the results from model 2 at two different dates, one on a particularly cloudy day (2018-01-29) and another on a day with scattered clouds (2018-02-01). We observe that the weather affects the performance dramatically, with the model predicting significantly worse on a day with scattered clouds due to the high volatility of PV output. Despite this, our model is still capable of producing $95\%$ credible intervals that capture the ground truth PVE values.
This results in large variability in the NLPD, as seen in Table~\ref{table:final-results}.

The above observation suggests that in order to improve predictions, we need to take into account exogeneous regressors, such as weather data and satellite images, which is beyond the scope of this work. Preliminary work using GPs to predict PV from satellite images was conducted in \cite{lawati2020short}, which could be useful if we can combine it with our model.
Taking into account the spatial correlations between different PV locations, as seen in Figure~\ref{fig:pv-stations}, may also help to improve predictions; however our initial experiments using spatio-temporal state-space GPs \cite{hamelijnck2021spatio} demonstrated otherwise (see Appendix~\ref{app:spatio-temporal-GP} for more details). We plan on investigating these directions further in future work.

\section{Conclusion}
In this exploratory work, we investigated the use of GPs as a tool for predicting solar PV with uncertainties. By recasting this process as a state-space model, inference using a large database of historical PV readings becomes numerically tractable. Moreover, the method allows us to incorporate new data on the fly, making it amenable to online data assimilation, which is key to a successful model in weather forecasting/nowcasting. While we only considered the PV 
timeseries as inputs to our model, in future, we aim to further investigate the inclusion of (1) spatial correlations between PV stations, and (2) exogenous regressors such as weather data and satellite images, within this framework to further improve predictions.

\begin{ack}
We would like to thank William Wilkinson and Ollie Hamelijnck for their generous help with the \texttt{BayesNewton} package, which we used in our experiments.
\end{ack}

\bibliography{biblio} 

\begin{thebibliography}{10}

\bibitem{abramowitz1964handbook}
Milton Abramowitz and Irene~A Stegun.
\newblock {\em Handbook of mathematical functions with formulas, graphs, and
  mathematical tables}.
\newblock US Government printing office, 1964.

\bibitem{bbc_article}
{BBC}.
\newblock London narrowly avoided post-heatwave blackout.
\newblock
  \url{https://www-bbc-co-uk.cdn.ampproject.org/c/s/www.bbc.co.uk/news/uk-england-london-62296443.amp},
  2022.

\bibitem{chang2020fast}
Paul~E Chang, William~J Wilkinson, Mohammad~Emtiyaz Khan, and Arno Solin.
\newblock Fast variational learning in state-space {G}aussian process models.
\newblock In {\em 2020 IEEE 30th International Workshop on Machine Learning for
  Signal Processing (MLSP)}, 2020.

\bibitem{garud2021review}
Kunal~Sandip Garud, Simon Jayaraj, and Moo-Yeon Lee.
\newblock A review on modeling of solar photovoltaic systems using artificial
  neural networks, fuzzy logic, genetic algorithm and hybrid models.
\newblock {\em International Journal of Energy Research}, 2021.

\bibitem{hamelijnck2021spatio}
Oliver Hamelijnck, William Wilkinson, Niki Loppi, Arno Solin, and Theodoros
  Damoulas.
\newblock Spatio-temporal variational {G}aussian processes.
\newblock {\em Advances in Neural Information Processing Systems}, 2021.

\bibitem{hartikainen2010kalman}
Jouni Hartikainen and Simo S{\"a}rkk{\"a}.
\newblock Kalman filtering and smoothing solutions to temporal {G}aussian
  process regression models.
\newblock In {\em IEEE International Workshop on Machine Learning for Signal
  Processing}. IEEE, 2010.

\bibitem{Hyndman2014Forecasting}
Rob~J Hyndman and George Athanasopoulos.
\newblock Forecasting: Principles and practice.
\newblock {\em Online at https://otexts.com/fpp3/}, 2014.

\bibitem{kamthe2022iterative}
Sanket Kamthe, So~Takao, Shakir Mohamed, and Marc Deisenroth.
\newblock Iterative state estimation in non-linear dynamical systems using
  approximate expectation propagation.
\newblock {\em Transactions on Machine Learning Research}, 2022.

\bibitem{khan2017conjugate}
Mohammad Khan and Wu~Lin.
\newblock Conjugate-computation variational inference: {C}onverting variational
  inference in non-conjugate models to inferences in conjugate models.
\newblock In {\em Artificial Intelligence and Statistics}. PMLR, 2017.

\bibitem{lawati2020short}
Yahya~Al Lawati, Jack Kelly, and Dan Stowell.
\newblock Short-term prediction of photovoltaic power generation using
  {G}aussian process regression.
\newblock {\em arXiv preprint arXiv:2010.02275}, 2020.

\bibitem{luo2021deep}
Xing Luo, Dongxiao Zhang, and Xu~Zhu.
\newblock Deep learning based forecasting of photovoltaic power generation by
  incorporating domain knowledge.
\newblock {\em Energy}, 2021.

\bibitem{lutkepohl2005new}
Helmut L{\"u}tkepohl.
\newblock {\em New introduction to multiple time series analysis}.
\newblock Springer Science \& Business Media, 2005.

\bibitem{ocf_nowcasting}
{Open Climate Fix}.
\newblock Nowcasting.
\newblock \url{https://www.openclimatefix.org/projects/nowcasting/}, 2022.

\bibitem{sarkka2013spatiotemporal}
Simo S{\"a}rkk{\"a}, Arno Solin, and Jouni Hartikainen.
\newblock Spatiotemporal learning via infinite-dimensional {B}ayesian filtering
  and smoothing: A look at {G}aussian process regression through {K}alman
  filtering.
\newblock {\em IEEE Signal Processing Magazine}, 2013.

\bibitem{solin2014explicit}
Arno Solin and Simo S{\"a}rkk{\"a}.
\newblock Explicit link between periodic covariance functions and state space
  models.
\newblock In {\em Artificial Intelligence and Statistics}. PMLR, 2014.

\bibitem{su2019machine}
Di~Su, Efstratios Batzelis, and Bikash Pal.
\newblock Machine learning algorithms in forecasting of photovoltaic power
  generation.
\newblock In {\em 2019 International Conference on Smart Energy Systems and
  Technologies (SEST)}. IEEE, 2019.

\bibitem{wilkinson2021bayes}
William~J Wilkinson, Simo S{\"a}rkk{\"a}, and Arno Solin.
\newblock Bayes-{N}ewton methods for approximate {B}ayesian inference with
  {PSD} guarantees.
\newblock {\em arXiv preprint arXiv:2111.01721}, 2021.

\bibitem{williams2006gaussian}
Christopher~KI Williams and Carl~Edward Rasmussen.
\newblock {\em Gaussian processes for machine learning}.
\newblock MIT press Cambridge, MA, 2006.

\bibitem{zhang2018deep}
Jinsong Zhang, Rodrigo Verschae, Shohei Nobuhara, and Jean-Fran{\c{c}}ois
  Lalonde.
\newblock Deep photovoltaic nowcasting.
\newblock {\em Solar Energy}, 2018.

\end{thebibliography}
\bibliographystyle{plain}

\appendix

\section{Background}\label{app:background}
In this Appendix, we provide background materials for the discussions in §\ref{sec:method}.

\subsection{Gaussian processes}
A {\em Gaussian process (GP)} on $\mathbb{R}^d$ is a random function $f : \mathbb{R}^d \rightarrow \mathbb{R}$ such that for any finite collection of points $\mat{X} = (x_1, \ldots, x_N) \in \mathbb{R}^d \times \cdots \times \mathbb{R}^d$, the random variable $\vec{f} := (f(x_1), \ldots, f(x_N)) \in \mathbb{R}^N$ is jointly Gaussian. They are uniquely characterised by a mean function $\mu : \mathbb{R}^d \rightarrow \mathbb{R}$ and a covariance kernel $k : \mathbb{R}^d \times \mathbb{R}^d \rightarrow \mathbb{R}$, where the former satisfies $\mu(x) = \mathbb{E}[f(x)]$ and the latter satisfies $k(x, x') = \mathrm{Cov}(f(x), f(x'))$ for all $x, x' \in \mathbb{R}^d$.

In statistical modelling, they can be used to specify priors over latent functions, whose distributions are updated as we acquire direct or indirect observations of this function. For example, when the dataset $\mathcal{D} = \{(x_i, y_i)\}_{i=1}^N$ consists of direct observations of $f$ corrupted by i.i.d. Gaussian noise, i.e., $y_i = f(x_i) + \epsilon_i$ where $\epsilon_i \sim \mathcal{N}(0, \sigma^2)$, we can compute our updated belief on $f$ as
\begin{align}
    \mathbb{E}[f(\cdot) | \vec{y}] &= \vec{k}_\mat{X}(\cdot)^T (\mat{K}_{\mat{X}\mat{X}} + \sigma^2 \mat{I})^{-1}\vec{y},  \label{eq:gp-posterior-mean}\\
    \mathrm{Cov}\left(f(\cdot) | \vec{y}\right) &= k(\cdot, \cdot) - \vec{k}_\mat{X}(\cdot)^T (\mat{K}_{\mat{X}\mat{X}} + \sigma^2 \mat{I})^{-1} \vec{k}_\mat{X}(\cdot). \label{eq:gp-posterior-cov}
\end{align}
Here, $\mat{K}_{\mat{X}\mat{X}} = k(\mat{X}, \mat{X}) \in \mathbb{R}^{N \times N}$ is the covariance matrix of $f$ evaluated at all the training inputs and $\vec{k}_\mat{X}(\cdot) \in \mathbb{R}^N$ is the cross-covariance of $f$ evaluated at the training inputs and at a query point.

\subsection{SDE representation of Gaussian processes} \label{app:sde-representation}
Computing the moment updates \eqref{eq:gp-posterior-mean}--\eqref{eq:gp-posterior-cov} pose several practical challenges that need to be considered. Firstly, we observe that the computation entails storing and inverting an $N \times N$ matrix $\mat{K}_{XX} + \sigma^2 \mat{I}$, which has $\mathcal{O}(N^2)$ memory and $\mathcal{O}(N^3)$ compute cost. When the dataset is large (for example in the order of millions), the problem becomes intractable. Secondly, equations \eqref{eq:gp-posterior-mean}--\eqref{eq:gp-posterior-cov} do not allow for efficient assimilation of new datapoints -- whenever a new datapoint $(x_*, y_*)$ is acquired, \eqref{eq:gp-posterior-mean}--\eqref{eq:gp-posterior-cov} has to be recomputed every time with the augmented dataset $\mathcal{D}' = \mathcal{D} \cup \{(x_*, y_*)\}$.

In \cite{hartikainen2010kalman}, a novel reformulation of GPs in one-dimension is presented, framing them as stochastic differential equations (SDEs). Specifically, given a zero-mean GP $f \sim \mathcal{GP}(0, k(\cdot, \cdot))$ defined over the real line, we can identify $f$ with an SDE
\begin{align} \label{eq:sde-form}
    \diff \vec{x}(t) = \mat{F} \vec{x}(t) \,\diff t + \mat{L} \,\diff \vec{W}_t, \quad \vec{x}(0) \sim \mathcal{N}(\vec{0}, \mat{P}_\infty),
\end{align}
where $\vec{x}(t)$ is an abstract state vector related to the original process $f(t)$ via $f(t) = \mat{H}\vec{x}(t)$ for some linear operator $\mat{H}$. Also, $\mat{F}, \mat{L} \in \mathbb{R}^{n \times n}$ are real matrices and $\vec{W}_t \in \mathbb{R}^{n}$ is a vector-valued Wiener process. The matrices $\mat{H}, \mat{F}, \mat{L}$ and the initial state covariance $\mat{P}_\infty$ are determined (up to approximations) by the choice of kernel $k$.
\begin{example}[Mat\'ern-3/2 GP]
We look at the SDE representation of the Matérn GP with $\nu = 3/2$, which can be found in \cite{hartikainen2010kalman, sarkka2013spatiotemporal}. This kernel has the form
\begin{align}
    k_{3/2}(t, t') = \sigma^2 \left(1+\frac{\sqrt{3}|t-t'|}{l}\right)\exp\left(-\frac{\sqrt{3}|t-t'|}{l}\right),
\end{align}
where $l$ is the lengthscale hyperparameter and $\sigma$ is the amplitude hyperparameter. By following the arguments in~\cite{hartikainen2010kalman}, we can represent it as the SDE
\begin{align}
    \diff
    \begin{bmatrix}
    f(t) \\
    f'(t)
    \end{bmatrix}
    =
    \begin{bmatrix}
    0 & 1 \\
    -\lambda^2 & -2\lambda
    \end{bmatrix}
    \begin{bmatrix}
    f(t) \\
    f'(t)
    \end{bmatrix}
    \diff t +
    \begin{bmatrix}
    0 & 0 \\
    0 & \sqrt{q}
    \end{bmatrix}
    \diff \mat{W}_t, \label{eq:matern-sde}
\end{align}
where $\lambda = \sqrt{3}/l$, $q = 4\sigma^2 \lambda^3$ and $f'(t)$ denotes the first derivative of $f(t)$. Moreover, the initial condition for the system is given by
\begin{align}
    \begin{bmatrix}
    f(0) \\
    f'(0)
    \end{bmatrix}
    \sim
    \mathcal{N}
    \left(
    \begin{bmatrix}
    0 \\
    0
    \end{bmatrix},
    \begin{bmatrix}
    \sigma^2 & 0 \\
    0 & -\sigma^2 \lambda^2
    \end{bmatrix}
    \right). \label{eq:matern-initial}
\end{align}
The solution $f(t)$ to the system~\eqref{eq:matern-sde}--\eqref{eq:matern-initial} is exactly equivalent to the Mat\'ern-3/2  GP in distribution, that is, $f(\cdot) \sim \mathcal{GP}(0, k_{3/2}(\cdot, \cdot))$. In short, the following matrices determine the state-space representation of the Mat\'ern-3/2 GP:
\begin{align}
    \mat{H} = [1, 0], \quad
    \mat{F} = \begin{bmatrix} 0 & 1 \\ -\lambda^2 & -2\lambda \end{bmatrix}, \quad
    \mat{L} = \begin{bmatrix} 0 & 0 \\ 0 & \sqrt{q} \end{bmatrix}, \quad
    \mat{P}_\infty = \begin{bmatrix} \sigma^2 & 0 \\ 0 & -\sigma^2 \lambda^2 \end{bmatrix}.
\end{align}
\end{example}

\begin{example}[Periodic GP]
The periodic kernel has the explicit form
\begin{align}
    k_{p}(t, t') = \sigma^2 \exp\left(-\frac{2\sin^2\left(\omega_0 \frac{t-t'}{2}\right)}{l^2}\right),
\end{align}
where $l$ is the lengthscale hyperparameter, $\sigma$ is the amplitude hyperparameter and $\omega_0$ is the frequency scale hyperparameter. In \cite{solin2014explicit}, it is shown that this kernel can be approximately expressed as a system of first-order differential equations
\begin{align}
    \frac{\diff x_j(t)}{\diff t} &= -j \omega_0 y_j(t) \label{eq:periodic-x} \\
    \frac{\diff y_j(t)}{\diff t} &= j \omega_0 x_j(t), \label{eq:periodic-y}
\end{align}
for $j = 0, \ldots, J$, where $J \in \mathbb{N}$ is the approximation order. As we can see, the stochasticity does not appear in the equations themselves, but only in the initial conditions
\begin{align}
    \begin{bmatrix}
    x_j(0) \\
    y_j(0)
    \end{bmatrix}
    \sim
    \mathcal{N}
    \left(\vec{0}, q_j^2 \mat{I} \right), \quad j = 0, \ldots, J, \label{eq:periodic-initial}
\end{align}
Here, the variance has the expression
\begin{align}
    q_j^2 = 
    \begin{cases}
    \frac{I_0(l^{-2})}{\exp(l^{-2})}, \quad \text{for } j = 0\\\\
    \frac{2I_j(l^{-2})}{\exp(l^{-2})}, \quad \text{otherwise}
    \end{cases},
\end{align}
where $I_j$ is the modified Bessel function of the first kind~\cite{abramowitz1964handbook}. From the system of random differential equations \eqref{eq:periodic-x}--\eqref{eq:periodic-initial}, we obtain an approximation to the periodic GP by $f(t) \approx \sum_{j=0}^J x_j(t)$. Putting this together, we see that the following matrices determine the state-space representation of the Periodic GP:
\begin{align}
    \mat{H} = [\mat{H}_1, \ldots, \mat{H}_J], \quad
    &\mat{F} = \mathtt{block\_diag}(\mat{F}_1, \ldots, \mat{F}_J), \\
    \mat{L} = \mathtt{block\_diag}(\mat{L}_1, \ldots, \mat{L}_J), \quad
    &\mat{P}_\infty = \mathtt{block\_diag}(\mat{P}_{\infty, 1}, \ldots, \mat{P}_{\infty, J}),
\end{align}
where
\begin{align}
    \mat{H}_j = [1, 0], \quad
    \mat{F}_j = \begin{bmatrix} 0 & -j\omega_0 \\ j\omega_0 & 0 \end{bmatrix}, \quad
    \mat{L}_j = \begin{bmatrix} 0 & 0 \\ 0 & 0 \end{bmatrix}, \quad
    \mat{P}_{\infty, j} = q_j^2 \mat{I}.
\end{align}
\end{example}

We can also combine two or more kernels together to form a new kernel, either by summing them or taking products. Below, we summarise the state-space representation of such kernels.

\begin{example}[Sum and product kernels]
Let $k_1(\cdot, \cdot)$ and $k_2(\cdot, \cdot)$ be two kernels, whose corresponding GPs have the SDE representation $\{(\mat{H}_i, \mat{F}_i, \mat{L}_i, \mat{P}_{\infty, i})\}_{i \in \{1,2\}}$.
The sum kernel $k(t, t') = k_1(t, t') + k_2(t, t')$ gives us a GP with the following SDE representation:
\begin{align}
    \mat{H} = [\mat{H}_1, \mat{H}_2], \quad
    \mat{F} = \mat{F}_1 \oplus \mat{F}_2, \quad
    \mat{L} = \mat{L}_1 \oplus \mat{L}_2, \quad
    \mat{P}_\infty = \mat{P}_{\infty, 1} \oplus \mat{P}_{\infty, 2},
\end{align}
where $\oplus$ denotes the direct sum $\mat{F}_1 \oplus \mat{F}_2 := \mathtt{block\_diag}(\mat{F}_1, \mat{F}_2)$. This amounts to solving the two SDEs independently and aggregating only at the end, i.e., $f(t) = \mat{H}_1 \vec{x}_1 + \mat{H}_2 \vec{x}_2 = f_1(t) + f_2(t)$.
The SDE representation of the product kernel $k(t, t') = k_1(t, t') \times k_2(t, t')$ on the other hand is given by \cite{solin2014explicit}:
\begin{align}
    \mat{H} = \mat{H}_2 \otimes \mat{H}_1, \quad
    \mat{F} = \mat{F}_2 \otimes \mat{I}_1 + \mat{I}_2 \otimes \mat{F}_1, \quad
    \mat{L} = \mat{L}_2 \otimes \mat{L}_1, \quad
    \mat{P}_\infty = \mat{P}_{\infty, 2} \otimes \mat{P}_{\infty, 1},
\end{align}
where $\otimes$ is the Kronecker product. This allows us to compute the SDE representation of a wide class of GPs, such as the quasi-periodic model \ref{eq:model-2} we considered in this work, from simple building blocks.
\end{example}

\subsection{Variational inference} \label{app:VI}
In cases where we have non-Gaussian likelihood,
direct inference becomes intractable. Hence, we resort to approximate inference methods. In this work, we consider the conjugate-computation variational inference (CVI) scheme \cite{khan2017conjugate}, which, as demonstrated in \cite{chang2020fast, hamelijnck2021spatio}, works well with spatio-temporal GPs, although we may have also used other inference methods such as approximate EP / power EP~\cite{kamthe2022iterative, wilkinson2021bayes}.

The goal of variational inference is to approximate an intractable posterior distribution $p(\vec{f} | \vec{y})$ by a tractable distribution $q(\vec{x})$, typically Gaussian with trainable mean and covariance, by minimising the KL-divergence $\mathcal{KL}(q(\cdot) \,||\, p(\cdot | \vec{y}))$. In CVI, this essentially boils down to finding an {\em approximate likelihood} $p(\vec{y} | \vec{f}) \approx \mathcal{N}(\widetilde{\vec{y}} | \vec{f}, \widetilde{\mat{R}})$ (note: this is seen as a function in $\vec{f}$, not $\vec{y}$!) such that
\begin{align} \label{eq:app-approx-posterior}
    q(\vec{f}) \propto \mathcal{N}(\widetilde{\vec{y}} | \vec{f}, \widetilde{\mat{R}}) p(\vec{f}),
\end{align}
where $p(\vec{f})$ is the GP prior. Denoting by $\widetilde{\vec{\lambda}} = (\widetilde{\vec{\lambda}}^{(1)}, \widetilde{\vec{\lambda}}^{(2)})$ the natural parameters of the approximate likelihood, that is, $\widetilde{\vec{\lambda}}^{(1)} = \widetilde{\mat{R}}^{-1} \widetilde{\vec{y}}$ and $\widetilde{\vec{\lambda}}^{(2)} = \widetilde{\mat{R}}^{-1}$, CVI proceeds via the following update rule \cite{hamelijnck2021spatio}:
\begin{align}\label{eq:app-cvi-update}
    \widetilde{\vec{\lambda}} \leftarrow (1-\beta)\widetilde{\vec{\lambda}} + \beta \frac{\partial \mathbb{E}_{q(\vec{f})}[\log p(\vec{y} | \vec{f})]}{\partial [\vec{m}, \vec{m}\vec{m}^T + \mat{P}]},
\end{align}
where $\vec{m}, \mat{P}$ are the current mean and covariance of the approximate posterior $q(\vec{f})$ and $\beta$ is the learning rate. In practice, we can compute the gradients in \eqref{eq:app-cvi-update} by using the Monte-Carlo method and the reparameterisation trick.
Converting this back to the moment parameterisation, we get
\begin{align}\label{eq:app-moment-parameterisation}
    \widetilde{\vec{y}}_{\text{new}} = (\widetilde{\vec{\lambda}}^{(2)}_{\text{new}})^{-1}\widetilde{\vec{\lambda}}^{(1)}_{\text{new}} \quad \text{and} \quad \widetilde{\mat{R}}_{\text{new}} = (\widetilde{\vec{\lambda}}^{(2)}_{\text{new}})^{-1}.
\end{align}
When the likelihood is separable, that is,
\begin{align}
    p(\vec{y} | \vec{f}) = \prod_{n=1}^N p(y_n | f_n),
\end{align}
we can apply updates on the marginals $q(f_n)$ to compute approximate likelihoods $p(y_n | f_n) \approx \mathcal{N}(\widetilde{y}_n | f_n, \widetilde{R}_n)$ at each time step independently, resulting in a diagonal $\widetilde{\mat{R}}$ matrix in \eqref{eq:app-approx-posterior}.

\subsection{State-space formulation}
Given the above discussions, we can now express our model
\begin{align} \label{eq:model}
    f \sim \mathcal{GP}(0, k(\cdot, \cdot)) \quad y_n \sim p(y_n | f(t_n)), \quad n = 1, \ldots, N,
\end{align}
in {\em state-space form} to perform inference using the Kalman filter. This reads
\begin{align}
    &\vec{x}_{n+1} = \mat{A}_n \vec{x}_n + \vec{\varepsilon}_n, \quad \vec{\varepsilon}_n \sim \mathcal{N}(\vec{0}, \mat{Q}_n), \label{eq:app-transition}\\
    &\widetilde{y}_{n+1} = \mat{H} \vec{x}_{n+1} + \eta_{n+1}, \quad \eta_{n+1} \sim \mathcal{N}(\vec 0, \widetilde{R}_{n+1}), \label{eq:app-measurement}
\end{align}
for $n = 0, 1, \ldots, N-1$, where $\mat{A}_n := \exp\left(\mat{F} \Delta t_n \right)$, $\mat{Q}_n := \mat{L} (\Delta t_n) \mat{L}^T$ (recall the matrices $\mat{F}, \mat{L}$ from §\ref{app:sde-representation}), $\widetilde{\vec{y}} = (\widetilde{y}_1, \ldots, \widetilde{y}_N), \widetilde{\mat{R}} = \mathtt{diag}(\widetilde{R}_1, \ldots, \widetilde{R}_N)$ are the mean and covariance respectively of the approximate likelihood computed in §\ref{app:VI}, and $\Delta t_n$ is taken to be the distance between two consecutive input data points $\Delta t_n = t_{n+1} - t_n$. The initial state distribution is $\vec{x}_0 \sim \mathcal{N}(\vec{0}, \mat{P}_\infty)$.

The full filtering algorithm is shown in Algorithm~\ref{alg:filter}. We denoted the mean and covariance of the filtering distribution $p(\vec{x}_n | \vec{y}_{1:n})$ by $\vec{m}_{n|n}, \mat{P}_{n|n}$, respectively, and likewise, the mean and covariance of the predictive distribution $p(\vec{x}_{n+h} | \vec{y}_{1:n})$ by $\vec{m}_{n+h|n}, \mat{P}_{n+h|n}$ for $h = 1, 2, \ldots$.
By using the Kalman filter, we can perform GP inference that only scales linearly in $N$ as opposed to the cubic cost in vanilla GP inference. Moreover, it has the added advantage of being able to assimilate new data on-the-fly, allowing for applications in online settings.

\begin{algorithm}[h]
	\caption{Filtering algorithm for model~\eqref{eq:model}}
	\label{alg:filter}
	\begin{algorithmic}[1]
		\State \textbf{Init:} $\vec{m}_{0|0} = \vec{0}$, $\mat{P}_{0|0} = \mat{P}_\infty$
			\For{$n
			= 0$ to $N-1$}
			\State 1. Prediction:
			\begin{align}
			    \vec{m}_{n+1|n} &= \mat{A}_n \vec{m}_{n|n} \\
			    \mat{P}_{n+1|n} &= \mat{A}_n \mat{P}_{n|n} \mat{A}_n^T + \mat{Q}_n
			\end{align}
			
			\While{Stopping criterion not met}
			\State 2. Update $(\widetilde{\vec{y}}_{n+1}, \widetilde{\mat{R}}_{n+1})$ using CVI~ \eqref{eq:app-cvi-update}--\eqref{eq:app-moment-parameterisation}.
			
			\State 3. Compute Kalman gain:
			\begin{align}
			    \mat{K}_{n+1} = \mat{P}_{n+1|n} \mat{H}^T (\mat{H} \mat{P}_{n+1|n} \mat{H}^T + \widetilde{\mat{R}}_{n+1})^{-1}
			\end{align}
			\State 4. Update moments:
			\begin{align}
			    \vec{m}_{n+1|n+1} &= \vec{m}_{n+1|n} + \mat{K}_{n+1}(\widetilde{\vec{y}}_{n+1} - \mat{H} \vec{m}_{n+1|n}) \\
			    \mat{P}_{n+1|n+1} &= (\mat{I} - \mat{K}_{n+1} \mat{H}) \mat{P}_{n+1|n} \\
			    q(f_{n+1}) &\leftarrow \mathcal{N}(f_{n+1} \,|\, \mat{H}\vec{m}_{n+1|n+1}, \mat{H}\mat{P}_{n+1|n+1}\mat{H}^T)
			\end{align}
			\EndWhile
		\EndFor
		
		\State \Return $(\vec{m}_{n|n}, \mat{P}_{n|n})_{n=0}^N$
	\end{algorithmic}
\end{algorithm}

\subsection{Hyperparameter optimisation}
We can also train our model on the dataset to find optimal hyperparameters $\mat{\Theta}$ using the CVI framework. The hyperparameters in this case include the GP hyperparemeters (such as the lengthscale $l$) and hyperparemeters in the likelihood distribution. This typically proceeds by maximizing the marginal likelihood
\begin{align}
    p(\vec{y}_{1:N} | \mat{\Theta}) = \int p(\vec{y}_{1:N} | \vec{f}, \mat{\Theta}) \,p(\vec{f} | \mat{\Theta}) \diff \vec{f},
\end{align}
which, for non-Gaussian $p(\vec{y}_{1:N} | \vec{f})$, does not have a closed form. Instead, what we do is to maximise the so-called evidence lower bound (ELBO)
\begin{align} \label{eq:app-elbo}
    \mathcal{L}_{\text{ELBO}}(\mat{\Theta}, q) = \mathbb{E}_{q(\vec{f})} \left[\log \frac{p(\vec{y} | \vec{f}, \mat{\Theta}) p(\vec{f}| \mat{\Theta})}{q(\vec{f})}\right],
\end{align}
for some tractable distribution $q$, which satisfies
\begin{align}
    \log p(\vec{y}_{1:N} | \mat{\Theta}) = \mathcal{L}_{\text{ELBO}}(\mat{\Theta}, q) + \mathcal{KL}(q(\vec{f}) \,||\, p(\vec{f} | \vec{y}, \mat{\Theta})).
\end{align}
Thus, for $q(\vec{f}) \approx p(\vec{f} | \vec{y}, \mat{\Theta})$, and therefore $\mathcal{KL}(q(\vec{f}) \,||\, p(\vec{f} | \vec{y}, \mat{\Theta})) \approx 0$, the ELBO acts as a lower-bound proxy to the log marginal-likelihood. To this end, we can do an EM-style update to find the optimal hyperparameters $\mat{\Theta}^*$, by alternating between updating the approximate posterior $q(\vec{f})$ (E-step) and doing a gradient-based maximisation of $\mathcal{L}_{\text{ELBO}}(\mat{\Theta}, q)$ with respect to $\mat{\Theta}$ for $q$ held fixed (M-step).

Assuming \eqref{eq:app-approx-posterior}, we can neatly perform this optimisation by combining it with the Kalman filter/smoother, as shown in \cite{hamelijnck2021spatio}.
In this case, the ELBO~\eqref{eq:app-elbo} can be re-expressed as:
\begin{align}
    \mathcal{L}_{\text{ELBO}}(\mat{\Theta}) &= \sum_{n=1}^N \mathbb{E}_{q(f_n)}\left[\log p(y_n | f_n, \mat{\Theta})\right] - \sum_{n=1}^N \mathbb{E}_{q(f_n)} \left[\log \mathcal{N}(\widetilde{y}_n | f_n, \widetilde{\mat{R}}_n)\right] \\
    &\quad + \sum_{n=1}^N \mathbb{E}_{p(\vec{x}_n | \vec{y}_{1:n-1})} \left[\mathcal{N}(\widetilde{y}_n | \mat{H} \vec{x}_n, \widetilde{\mat{R}}_n)\right].
\end{align}

The E-step in this setting becomes equivalent to updating the approximate marginal posteriors $\{q(f_n)\}_{n=1}^N$ so that $q(f_n) \approx p(f_n | \vec{y}_{1:N}, \mat{\Theta})$, which we can achieve via Gaussian smoothing (Algorithm~\ref{alg:smoothing}). We can also approximate the predictive distribution $p(\vec{x}_n | \vec{y}_{1:n-1})$ in the third term by $\mathcal{N}(\vec{x}_n \,|\, \vec{m}_{n|n-1}, \vec{P}_{n|n-1})$ using the filtering algorithm (Algorithm~\ref{alg:filter}).


\begin{algorithm}[h]
	\caption{Smoothing algorithm for model~\eqref{eq:model}}
	\label{alg:smoothing}
	\begin{algorithmic}[1]
		\State \textbf{Init:} Run Algorithm~\ref{alg:filter} to compute the filtering distributions $(\vec{m}_{n|n}, \mat{P}_{n|n})$ for $n=0, \ldots, N$
			\For{$n
			= N$ to $1$}
			\State \textbf{1. Prediction:}
			\begin{align}
			    \vec{m}_{n|n-1} &= \mat{A}_{n-1} \vec{m}_{n-1 | n-1} \\
			    \mat{P}_{n|n-1} &= \mat{A}_{n-1}^T \mat{P}_{n-1|n-1} \mat{A}_{n-1} + \mat{Q}_{n-1}
			\end{align}
			
			\State \textbf{2. Compute smoother gain:}
			\begin{align}
			    \mat{J}_{n-1} = \mat{P}_{n-1|n-1} \mat{A}_{n-1} \mat{P}_{n | n-1}^{-1}
			\end{align}
			\State \textbf{3. Update:}
			\begin{align}
			    \vec{m}_{n-1|N} &= \vec{m}_{n-1|n-1} + \mat{J}_{n-1}(\vec{m}_{n|N} - \vec{m}_{n|n-1}) \\
			    \mat{P}_{n-1|N} &= \mat{P}_{n-1|n-1} + \mat{J}_{n-1} (\mat{P}_{n|N} - \mat{P}_{n|n-1}) \mat{J}_{n-1}^T \\
			    q(f_{n-1}) &\leftarrow \mathcal{N}(f_{n-1} \,|\, \mat{H}\vec{m}_{n-1|N}, \mat{H}\mat{P}_{n-1|N}\mat{H}^T)
			\end{align}
		\EndFor
		
		\State \Return $(\vec{m}_{n|N}, \mat{P}_{n|N})_{n=0}^N$
	\end{algorithmic}
\end{algorithm}

\section{Experimental details}
Here, we provide further details on the experiments conducted in Section~\ref{sec:results}.

\subsection{Baselines} \label{app:baselines}
We consider a variety of baselines to assess the performance of the GP models applied to the PV nowcasting task.

\paragraph{Persistence.} The most simple and trivial benchmark is the persistence model, defined as:
\begin{align}
    \hat{y}_{n + 1} = y_n.
\end{align}
In other words, it predicts the next observations as the most recent value of the time series. This simple model is surprisingly hard to beat, and is a good first benchmark to compare against.

\paragraph{Yesterday.} A slight variation of the persitence is the yesterday model, defined as:
\begin{align}
    \hat{y}_{n + 1} = y_{n + 1 -  \#24\text{hr}},
\end{align}
where `\#24hr' indicates the number of timesteps corresponding to 24 hours in physical time.
This is a good baseline for data that shows strong daily seasonality.

\paragraph{Hourly smoothing.}
This is the first model here that performs any degree of smoothing (i.e., averaging of past values). This is given by (assuming that the time interval between each observations is 5 minutes):
\begin{align}
    \hat{y}_{n + 1} = \frac{1}{12} \sum_{i=1}^{12} y_{n + 1 - i}.
\end{align}
We take averages of the past $12$ data points since 1 hour = 5 minutes $\times$ 12 and we are smoothing over data in the past hour, as the name suggests.

\paragraph{Simple Exponential Smoothing.}
The simple exponential smoothing (EST) is one of the most effective methods of forecasting when no clear trend or seasonality are present \cite{Hyndman2014Forecasting}. This is similar to hourly smoothing, except that instead of cutting off the averaging at 1 hour and using uniform weights, the weights decay exponentially with time:
\begin{align}
    \hat{y}_{n + 1} = \sum_{j=0}^{J-1} \alpha(1-\alpha)^j y_{n-j} + (1-\alpha)^n \ell_0,
\end{align}
where $\alpha \in [0, 1]$ is the smoothing exponent, and $\ell_0$ is the initial value, also treated as a hyperparameter. We can also write this in ``component form" as:
\begin{align}
\begin{cases}
    \hat{y}_{n + 1} = \ell_n, \\
    \ell_n = \alpha y_n + (1-\alpha) \ell_{n-1}.
\end{cases}
\end{align}
Defining the ``error" $\varepsilon_n := y_n - \ell_{n-1}$, which we assume to be distributed according to $\mathcal{N}(0, \sigma^2)$ (again, $\sigma$ is a hyperparameter to be optimised), we can further write this in state-space form:
\begin{align}
\begin{cases}
    y_{n} = \ell_{n-1} + \varepsilon_{n}, \\
    \ell_n = \ell_{n-1} + \alpha \varepsilon_n.
\end{cases}
\end{align}
This is now a stochastic model, which allows us to get uncertainty estimates alongside predictions by means of Kalman filtering/smoothing.

\paragraph{Seasonal Exponential Smoothing.}
The seasonal EST method, also called Holt-Winters’ seasonal method \cite{Hyndman2014Forecasting}, is an extension of the EST algorithm to capture trend and seasonality in data. The additive model is defined by three smoothing equations: one for the level term $\ell_n$, one for the trend term $b_n$, and one for the seasonal term $s_n$, with parameters $\alpha$, $\beta^*$, and $\gamma$ respectively. Letting $m$ be the period (in timesteps) of said seasonality, the model is given by:
\begin{align}
\begin{cases}
    \hat{y}_{n + 1} = \ell_n + b_n + s_{n+1-m} \\
    \ell_n = \alpha(y_n - s_{n-m}) + (1-\alpha)(\ell_{n-1} + b_{n-1}) \\
    b_n = \beta^*(\ell_n - \ell_{n-1}) + (1-\beta^*) b_{n-1} \\
    s_n = \gamma(y_n - \ell_{n-1} - b_{n-1}) + (1-\gamma) s_{n-m}.
\end{cases}
\end{align}
Similarly as before, defining the error $\varepsilon_n = y_n - \ell_{n-1} - b_{n-1} \sim \mathcal{N}(0, \sigma^2)$, we can express this in state-space form:
\begin{align}
\begin{cases}
    y_{n} = \ell_{n-1} + b_{n-1} + s_{n-m} + \varepsilon_{n} \\
    \ell_n = \ell_{n-1} + b_{n-1} + \alpha \varepsilon_n \\
    b_n = b_{n-1} + \beta \varepsilon_n \\
    s_n = s_{n-m} + \gamma \varepsilon_n,
\end{cases}
\end{align}
where $\beta := \alpha \beta^*$. Again, this enables us to get uncertainty estimates via Kalman filtering/smoothing.

\paragraph{Vector Autoregression.}
The Vector Autoregressive model (VAR) is the only baseline that we considered capable of learning correlations between time series at different PV stations  \cite{lutkepohl2005new}. With VAR, the spatio-temporal process is modelled as a multivariate time series $\mat{Y} = (\vec{y}_1, \ldots, \vec{y}_{n_t}) \in \mathbb{R}^{n_t \times n_s}$, given by:
\begin{align}
    \vec{y}_n &= \vec{\nu} + \sum_{j=1}^J \mat{A}_j \vec{y}_{n-j} + \vec{\varepsilon}_t, \quad \vec{\varepsilon}_n \sim \mathcal{N}(0, \mat{\Sigma}),
\end{align}
where $\vec{\nu}$ is a vector of intercepts to allow for non-zero means, $J$ is the order of the lag, $(\mat{A}_i)_{i=1}^J$ are ($n_s \times n_s$) matrices, and $\mat{\Sigma} \in \mathbb{R}^{n_s \times n_s}$ is the error covariance. Since this is a stochastic model, we can also get uncertainty estimates via Kalman filtering/smoothing.

We implemented the baselines using the \texttt{statsmodels} package and selected the hyperparameters using Bayesian optimisation.

\subsection{Evaluation metrics}\label{app:evaluation-metric}
We evaluate the performance of our models on a hold-out set $\mathcal{D}^* = \{(t_i^*, y_i^*)\}_{i=1}^{N^*}$, by comparing the mean absolute error (MAE) and the negative log predictive density (NLPD), respectively, defined as:
\begin{align}
    \mathrm{MAE} &:=  \frac{1}{N^*} \sum_{i=1}^{N^*} \big|y_i^* - \mathbb{E}[f(t_i^*) \,|\, \mathcal{D}] \,\big|, \label{eq:app-mae} \\
    \mathrm{NLPD} &:= - \sum_{i=1}^{N^*} \mathrm{log} \,p(y_i^* | \mathcal{D}) \\
    &= -\sum_{i=1}^{N^*} \mathrm{log} \int p(y_i^* | \vec{f}^*)p(\vec{f}^* | \mathcal{D}) \diff \vec{f}^*, \label{eq:app-nll}
\end{align}
where $f$ is our model, $\mathcal{D}$ is the training set, and $\vec{f}^*$ denotes the vector of $f$ evaluated at the test input locations $(t_1^*, \ldots, t_{N^*}^*)$.
The former assesses only the quality of the predictive mean, while the latter also assesses the quality of the predictive uncertainty. In practice, we approximate the integral in \eqref{eq:app-nll} using Monte-Carlo method when we have a non-conjugate likelihood.

When the model $f$ is deterministic, we only evaluate the MAE, replacing the conditional mean $\mathbb{E}[f(t_i^*) \,|\, \mathcal{D}]$ in \eqref{eq:app-mae} by $f(t_i^*)$.

\subsection{Cross-validation method}\label{app:cross-validation}
\begin{figure}[h]
    \centering
    \includegraphics[width=4cm]{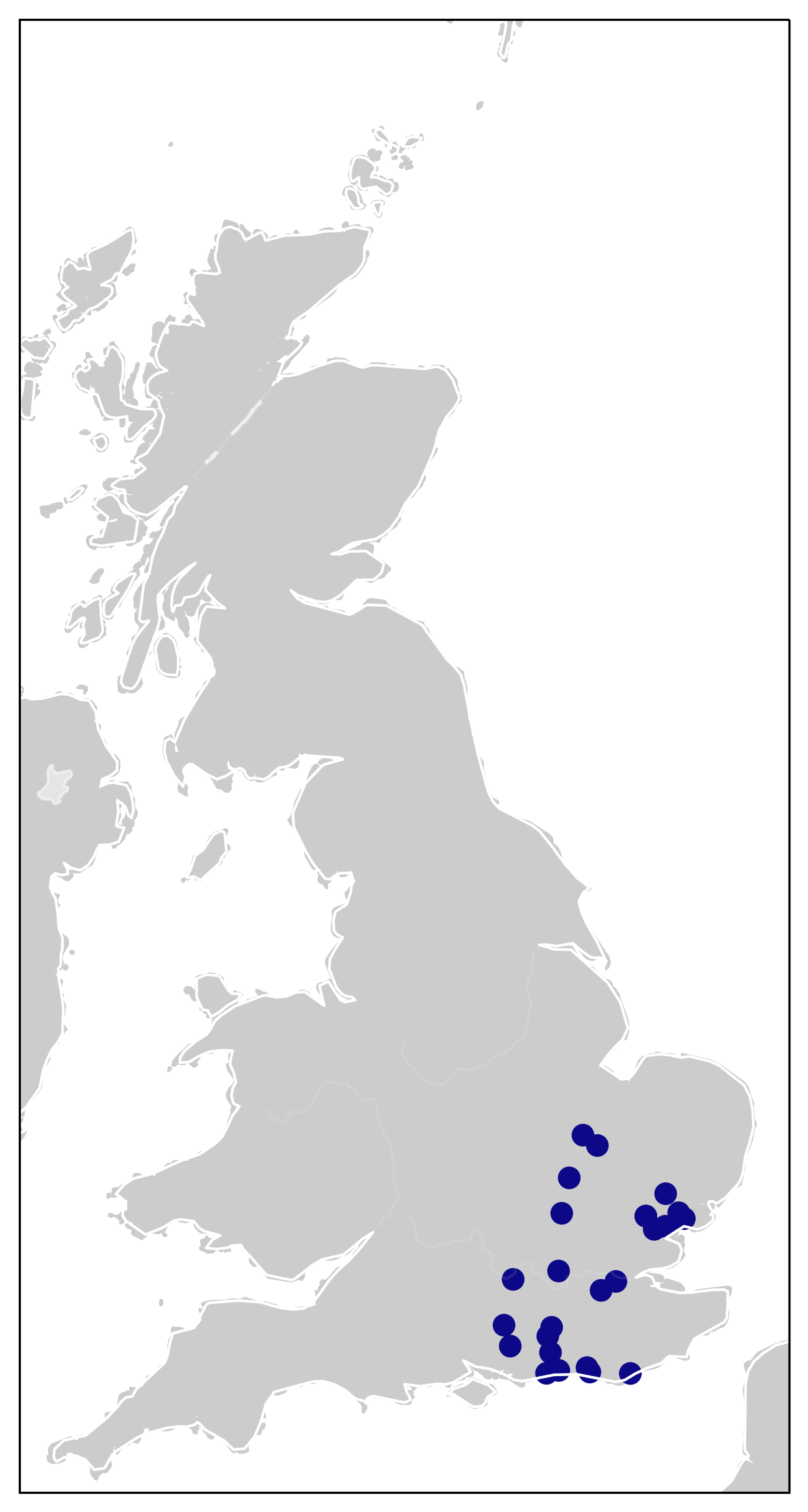}
    \caption[PVE systems in south-east England]{Plot of PV systems scattered in the south-east of England. This sub-sample of PV systems was used during the experimentation phase to be able to run more experiments with lower computational requirements.}
    \label{fig:london_sys}
\end{figure}

\begin{figure}[h]
    \centering
    \includegraphics[width=0.8\textwidth]{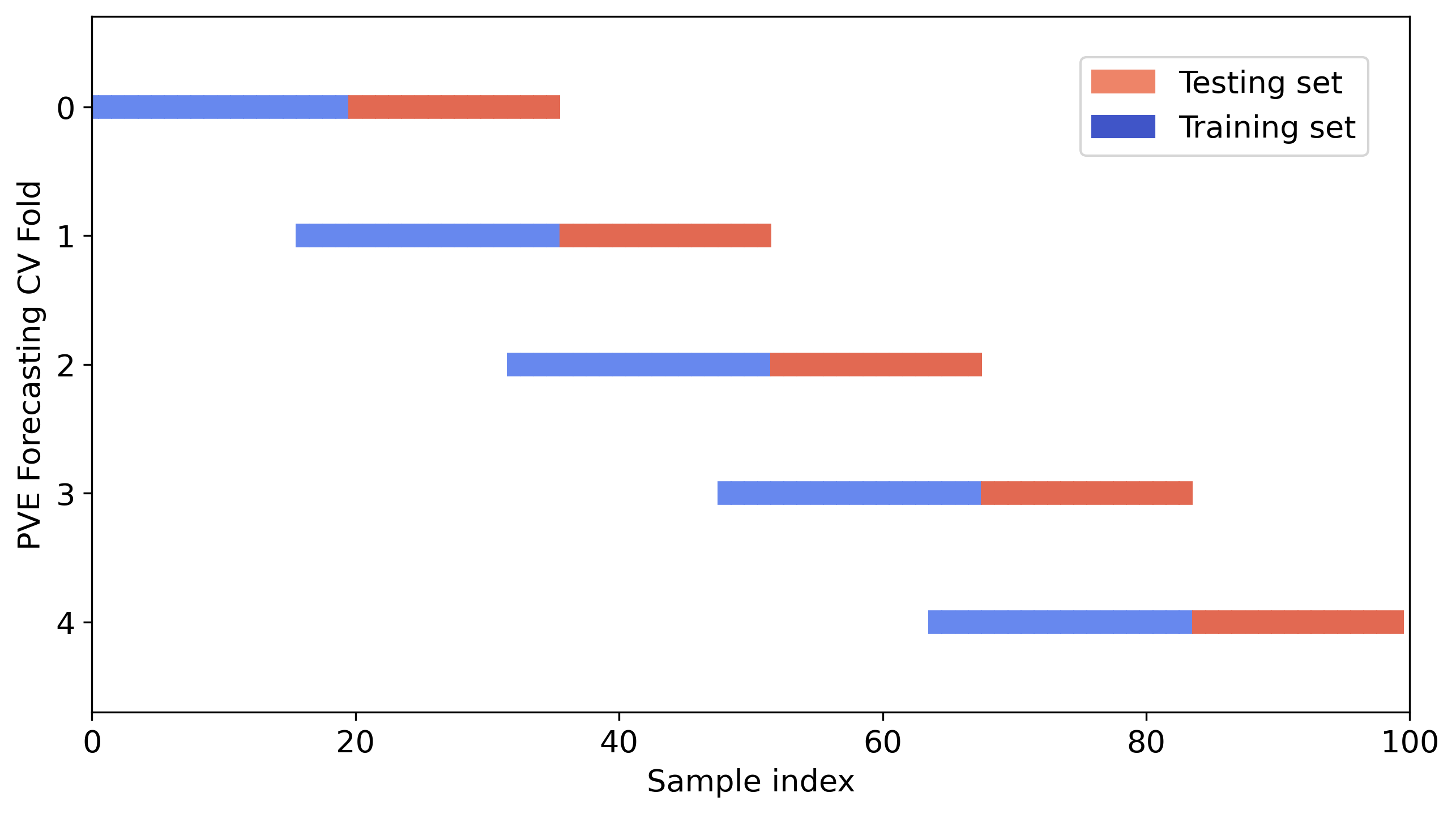}
    \caption{Illustration of our cross-validation scheme with rolling windows.}
    \label{fig:cross-validation}
\end{figure}

We evaluated our models on 27 PV systems located in the South-East of England, which we plot in Figure~\ref{fig:london_sys}, and used a walk-forward Cross-Validation (CV) with rolling window to assess their performance. Walk-forward CV must be used in time series data to avoid data leakage, as it ensures that the model never uses data that follows chronologically after the prediction window. In Figure~\ref{fig:cross-validation} we illustrate a typical time series split used in walk-forward CV.

In our CV scheme, we used training windows of 100 days and test windows of 2 hours. The models were trained and tested for 78 CV folds, each fold shifted by a day, but ensuring that the time of day at which the forecasts were executed differed between each fold. We also set our folds so that the forecast time always fell between 10:00 and 14:00, so that there are at least two hours of observations within the same day both before and after the forecast (recall that in our preprocessing step, we sliced the timeseries to include only data between 8:00 and 16:00). This was to ensure that the models were tested only on data within the same day to avoid discontinuity in the prediction time window.

To speed up the computation, we also initialised trainable parameters in each fold based on computations in the previous fold. In particular, model hyperparameters in each fold (except the first) were initialised with the final value in the previous fold and the variational parameters $(\widetilde{\vec{y}}, \widetilde{\mat{R}})$ in the approximate likelihood (see \ref{app:VI}) were reused at coinciding points in the training windows.

\section{Results from spatio-temporal GPs}\label{app:spatio-temporal-GP}

We have also done some initial experiments using spatio-temporal state-space GPs \cite{sarkka2013spatiotemporal}, to take into account spatial correlations between timeseries at different PV stations. We used separable kernels to model the spatio-temporal data, i.e.,
\begin{align}
    k((\vec{s}, t), (\vec{s}', t')) = k_s(\vec{s}, \vec{s}') \,k_t(t, t'),
\end{align}
where the same temporal kernels were used from §\ref{sec:model} and the Mat\'ern-3/2 kernel was used for the spatial component. Following \cite{hamelijnck2021spatio}, the same inference method that we described in Appendix \ref{app:background} hold for such kernels with slight modification, only that we get a coupled system of 27 stochastic differential equations, instead of 27 independent ones.

In Table~\ref{table:spatiotemporal_results}, we demonstrate the results of spatio-temporal GPs, one that is coupled with a simple Mat\'ern temporal component, which we called `model 1' in §\ref{sec:model}, and the other that is coupled with a quasi-periodic temporal kernel, which we called `model 2' in §\ref{sec:model}.

\begin{table}[H]
\centering
\begin{tabular}{|c|c|c|}
\hline
Model & MAE $\downarrow$ (mean $\pm$ std) & NLPD  $\downarrow$ (median $\pm$ m.a.d.) \\
\hline
Temporal Simple GP & $0.134 \pm 0.056$ & $-4.93 \pm 16.6$  \\ 
Temporal Quasi-Periodic GP & $\mathbf{0.109 \pm 0.050}$ & $\mathbf{-12.9 \pm 13.8}$ \\ 
\hline
Spatio-temporal Simple GP & $0.175 \pm 0.056$ & $-6.88 \pm 13.0$  \\
Spatio-temporal Quasi-Periodic GP & $0.140 \pm 0.058$ & $-6.89 \pm 29.9$ \\
\hline
\end{tabular}
\caption{Comparison of purely temporal GPs vs. spatio-temporal GPs. Our initial results suggest that including the spatial component deteriorates results rather than improving it, both in terms of the MAE and the NLPD.}
\label{table:spatiotemporal_results}
\end{table}

The results in Table~\ref{table:spatiotemporal_results} suggest that the spatial correlation between stations does not help improve our predictions as we had hoped, but rather deteriorates it. A possible reason for this is simply that we could not train our models until convergence as a result of the increased computational cost. It could also be suggesting that since we are predicting solar power directly, which can be affected by external factors such as the make of solar panels, orientation of solar panels, etc., it is better to model each timeseries individually. However, at first glance, there appears to be substantial spatial correlations between neighbouring systems, as we can see from Figure~\ref{fig:pv-stations}, so we may be able to improve on this upon further work.

\section{Additional plots}

\subsection{Exploratory data analysis}
Here, we plot the results of our exploratory analysis on the PV timeseries data.

In our exploratory analysis, we first studied the seasonality and autocorrelation properties of the PV timeseries, as seen for example in Figure~\ref{fig:example-timeseries}. A random PV system was selected for the analysis, however every timeseries showed very similar characteristics.

\begin{figure}[h]
    \centering
    \begin{subfigure}[b]{0.48\textwidth}
         \centering
         \includegraphics[width=\textwidth]{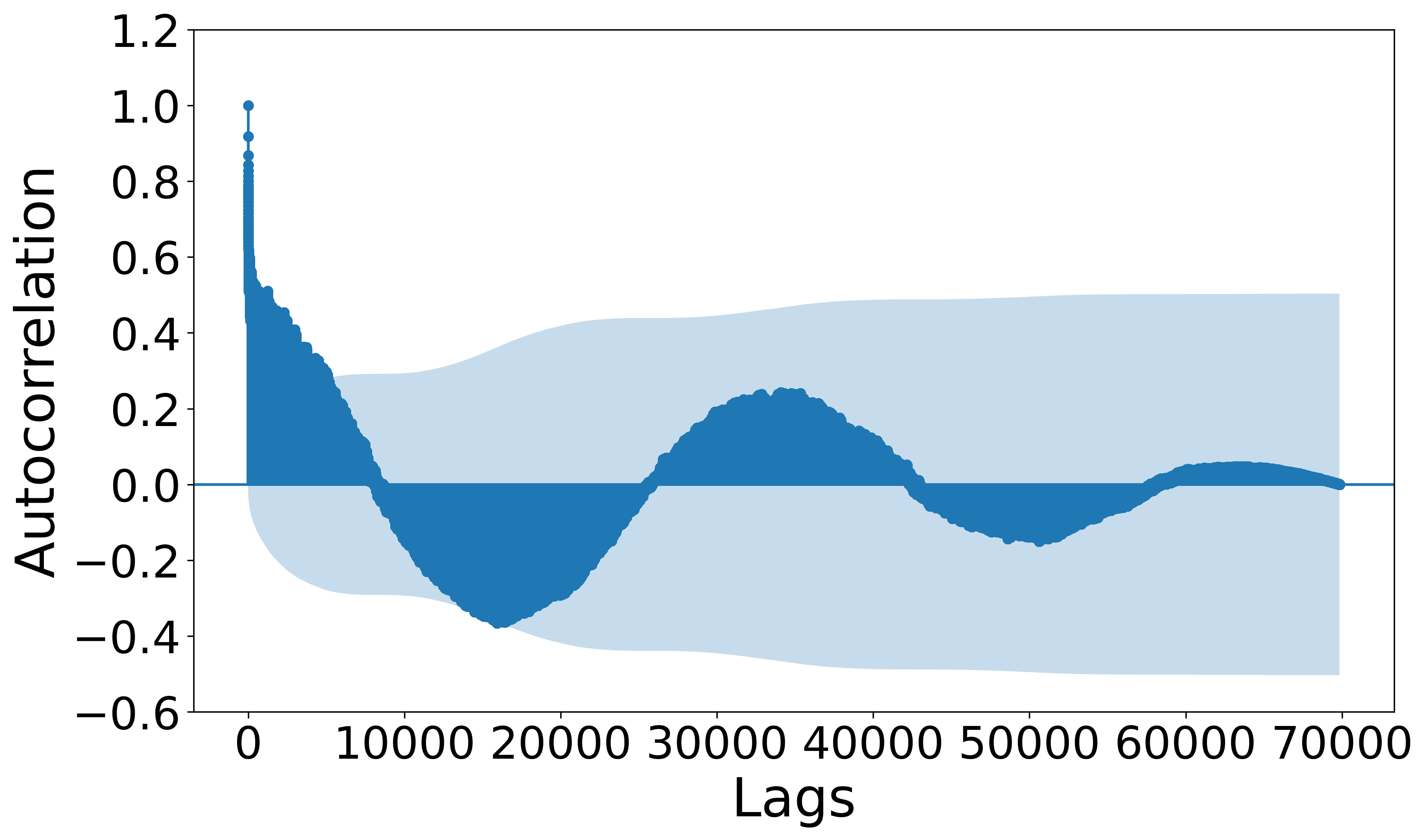}
         \caption{ACF for two-year window}
     \end{subfigure}
     \hfill
     \begin{subfigure}[b]{0.48\textwidth}
         \centering
         \includegraphics[width=\textwidth]{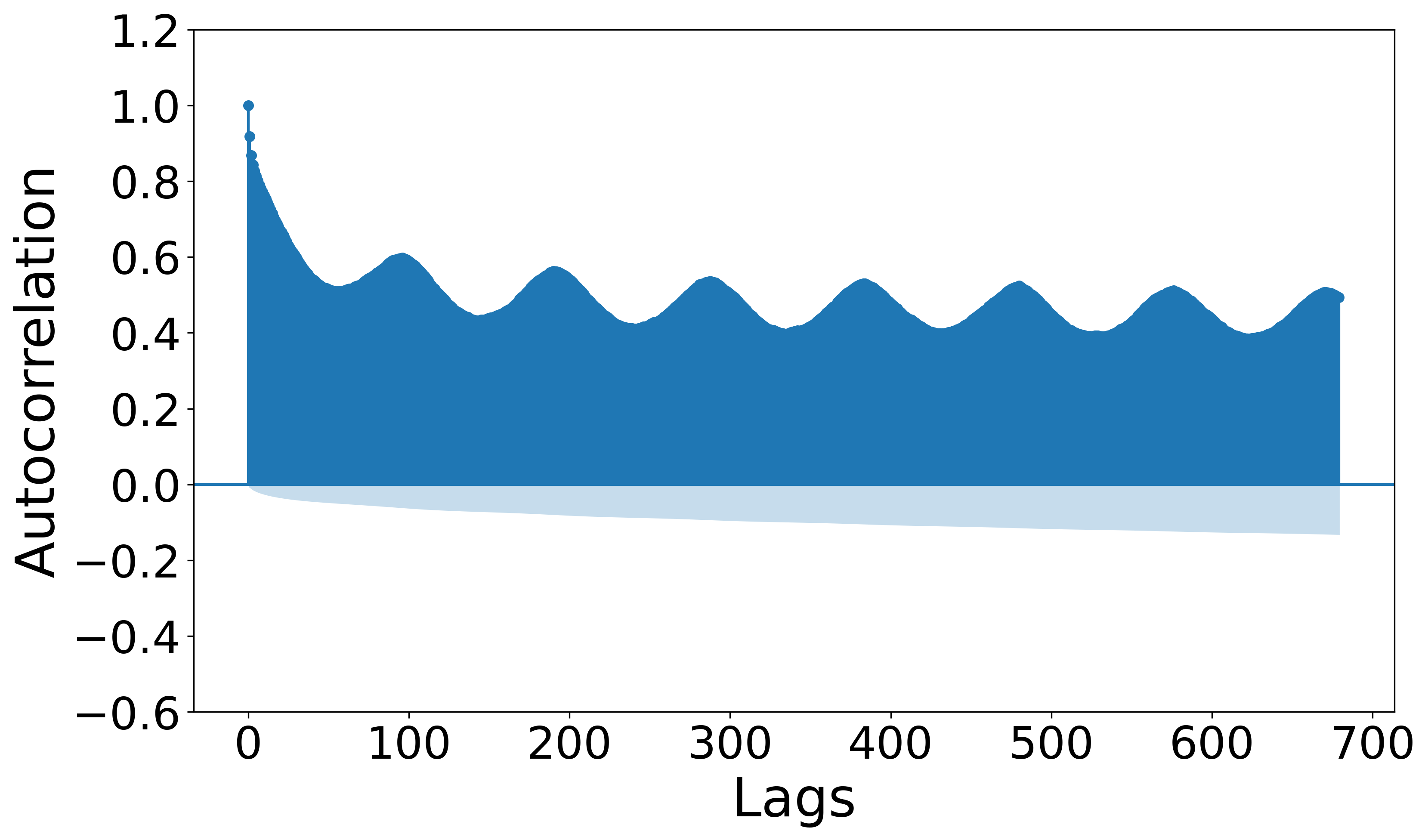}
         \caption{ACF for one-week window}
     \end{subfigure}
    \caption{ACF plots at windows of (a) two years, and (b) one-week. A clear daily and yearly seasonality is seen.}
    \label{fig:timeseries-autocorrelation}
\end{figure}

We used the autocorrelation function (ACF) to detect seasonality in our timeseries. The ACF is defined as
$\rho(k) = \frac{\gamma(k)}{\gamma(0)}$ where $\gamma(k) = \text{Cov}(y_i,y_i+k)$ is the autocovariance, and $\gamma(0)$ is the variance of the timeseries. An ACF close to 1 or -1 represents a high degree of autocorrelation, which can help identify seasonality in data. The ACF is shown in Figure~\ref{fig:timeseries-autocorrelation} for a two-year window and a one-week window respectively, where 96 lags correspond to one day of observations.

From the above plot, we see a clear daily and yearly seasonality. Moreover, the high values observed in the ACF at the first lag suggests high correlation with the most recent observation.

\begin{figure}[h]
    \centering
    \includegraphics[width=0.8\textwidth]{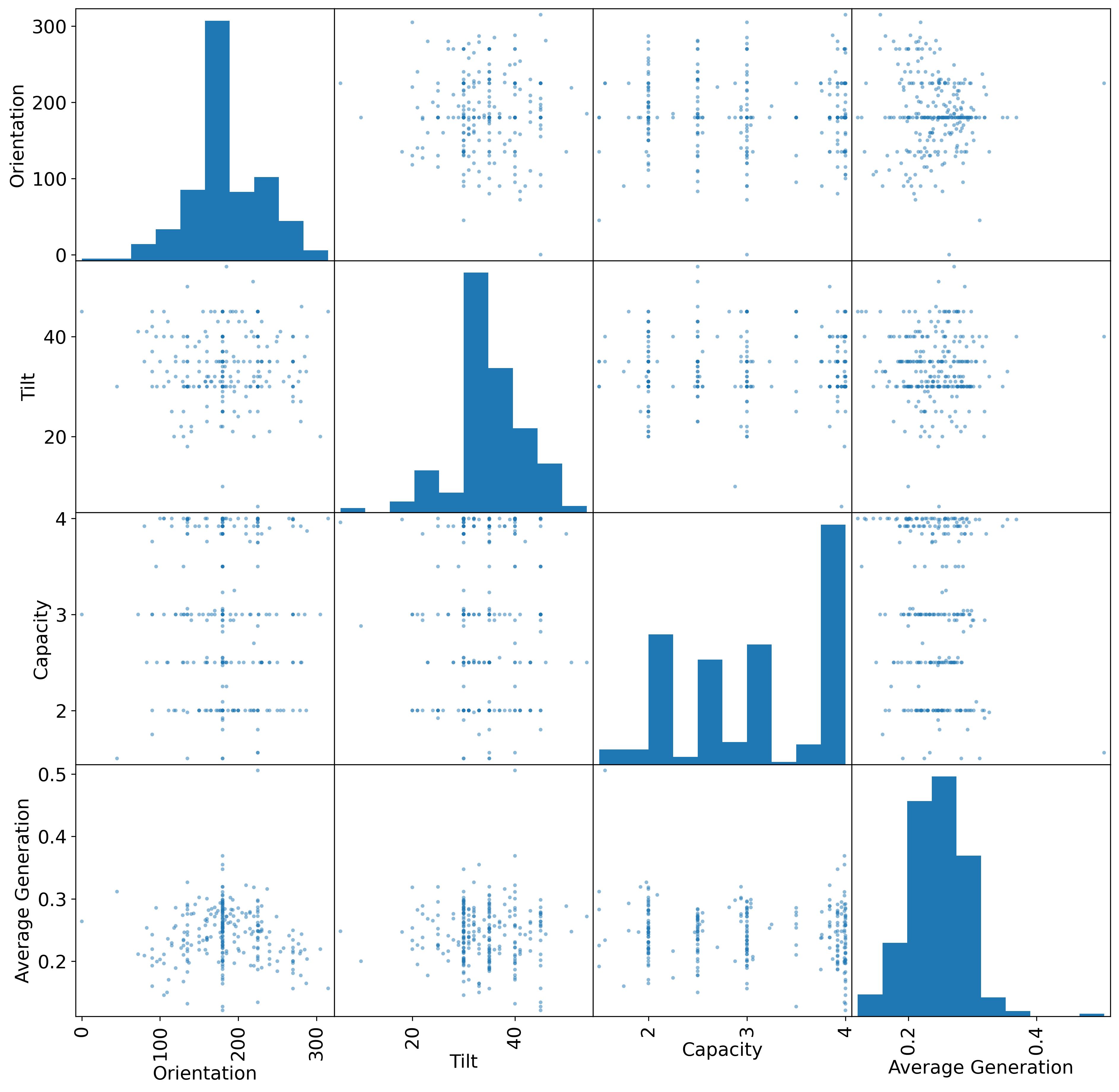}
    \caption{Scatter matrix of three explanatory variables - orientation, tilt, and capacity - and the average generation. No strong relationship can be observed between the explanatory variables and average PV generation.}
    \label{fig:app-feature-correlations}
\end{figure}

In the second step of our analysis, we studied the correlation between the PV values and available information on each PV station, including orientation, tilt, and capacity of each system. Intuitively, a system that is oriented and tilted towards the sun could yield more power than a system that is oriented and tilted away from the sun. Capacity was also studied to understand if there is any effect related to the size of the system and its expected production. The PV outputs were averaged over the entire time frame for each system. We plot the correlation matrices in Figure~\ref{fig:app-feature-correlations}.

The correlation matrices do not indicate any relationship between the three explanatory variables and the average PV generation.
In an effort to further identify any such relationship, an ordinary least square regression model was fitted, however none of the three explanatory variables were found to have a significant relationship with the response variable. For this reason, we deduce that none of the available explanatory system-specific variables will be useful for predicting PV output.

\subsection{Additional predictions}
\begin{figure*}[tb]
    \begin{minipage}{.48\textwidth}
        \centering
        \includegraphics[height=4cm]{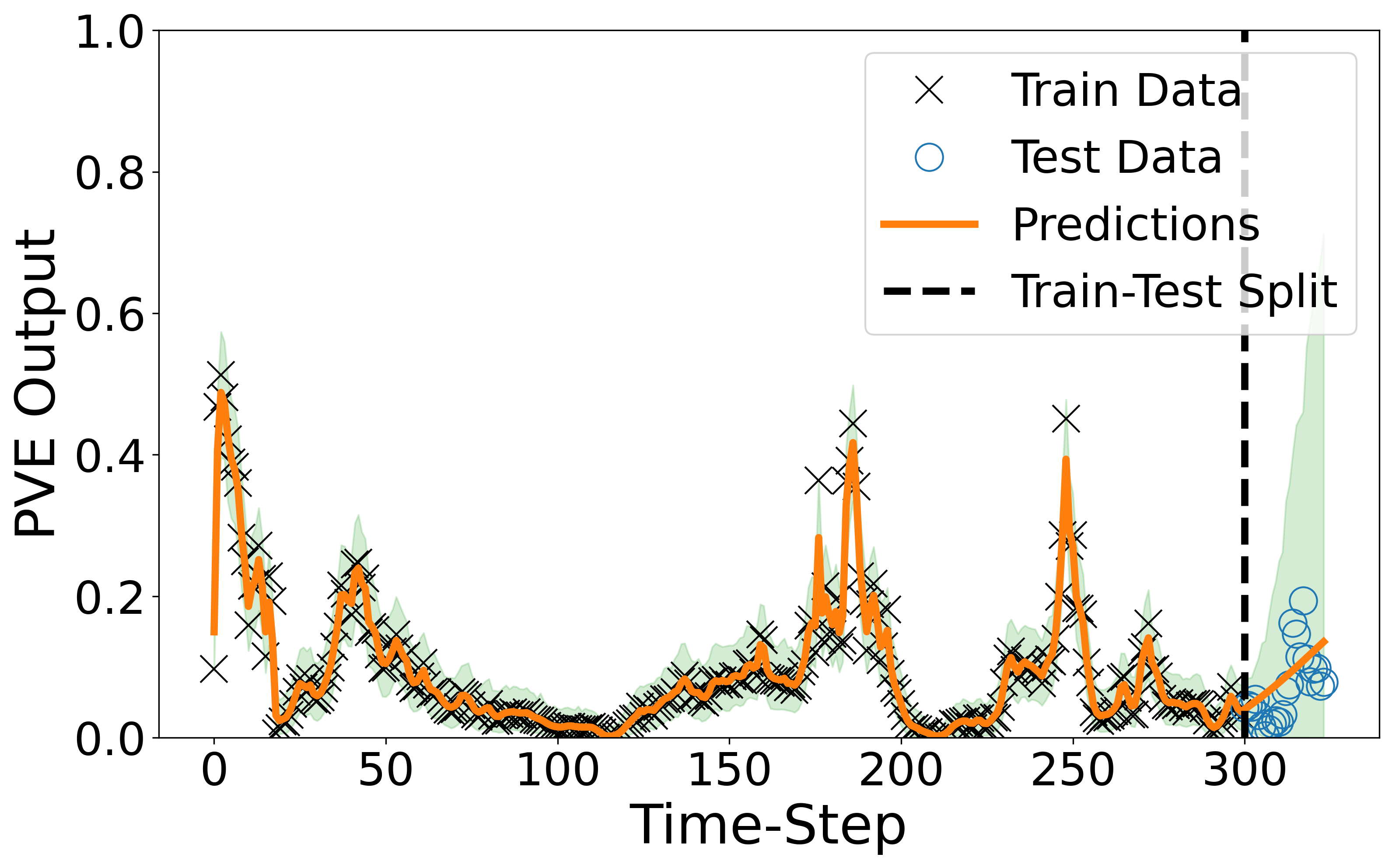}
        \par
        \vspace{15pt}
        \includegraphics[height=4cm]{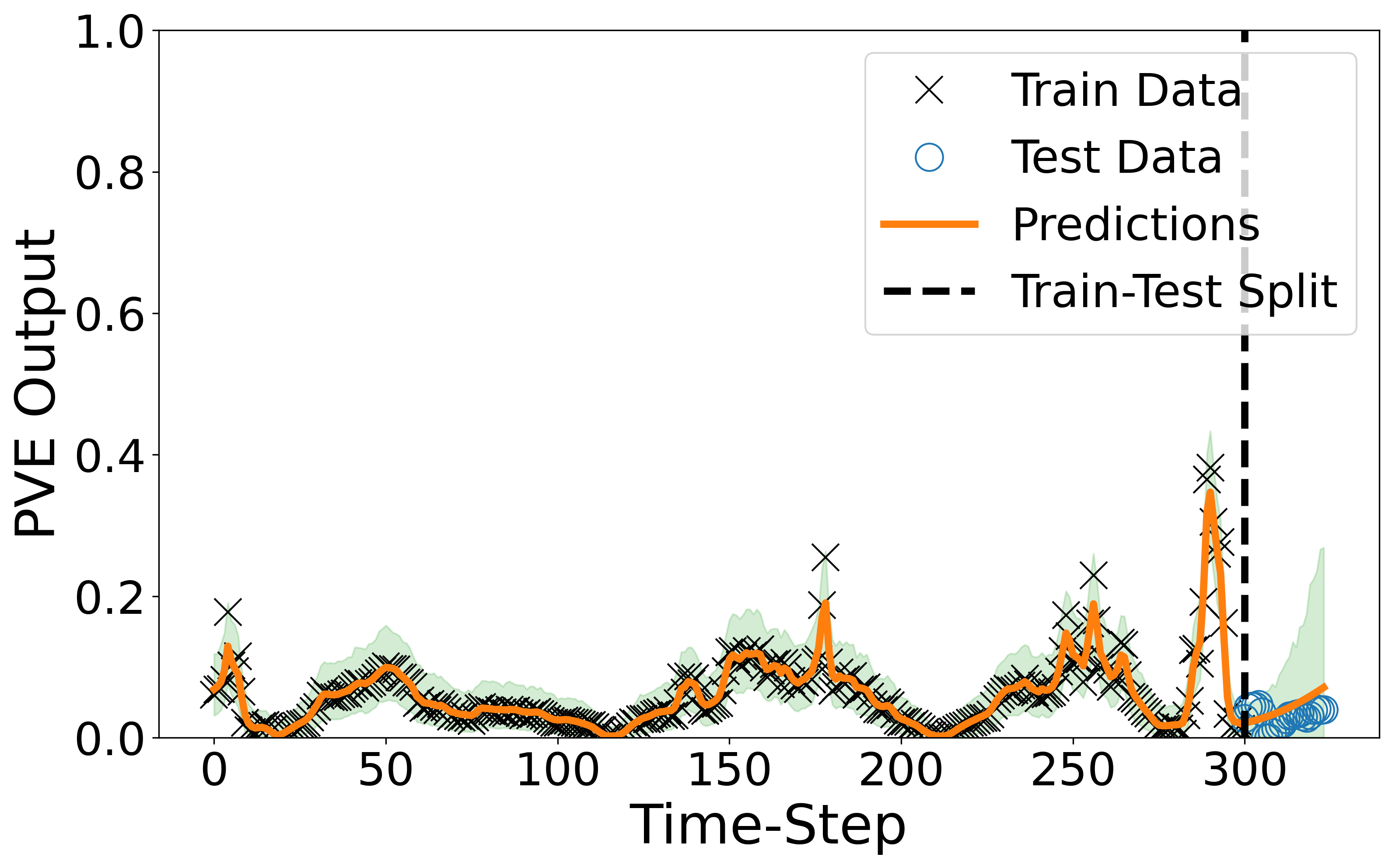}
        \par
        \vspace{15pt}
        \subfloat[Predictions at 2018-01-29]{\includegraphics[height=4cm]{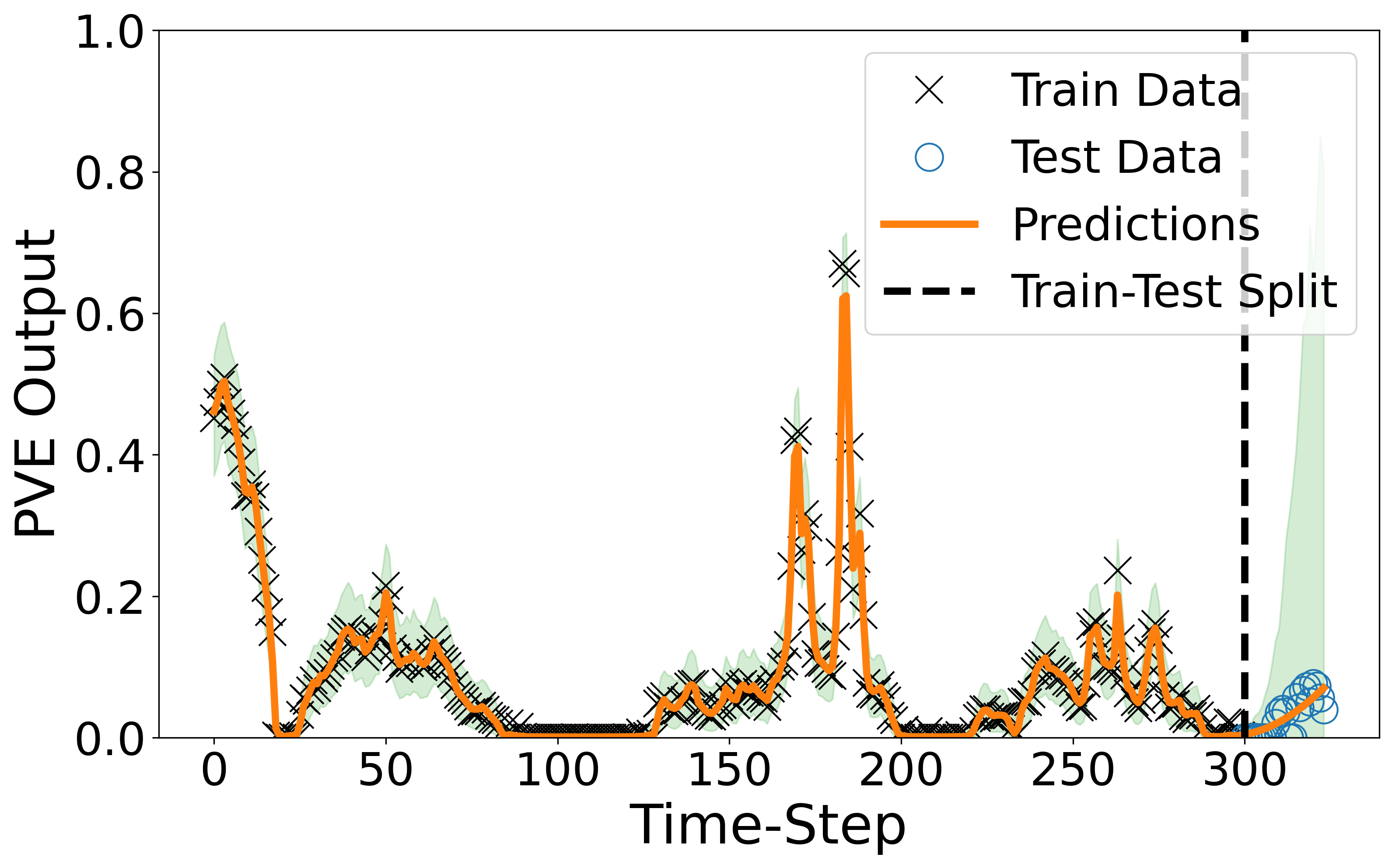}}
        \vspace{15pt}
        \subfloat[Predictions at 2018-01-29]{\includegraphics[height=4cm]{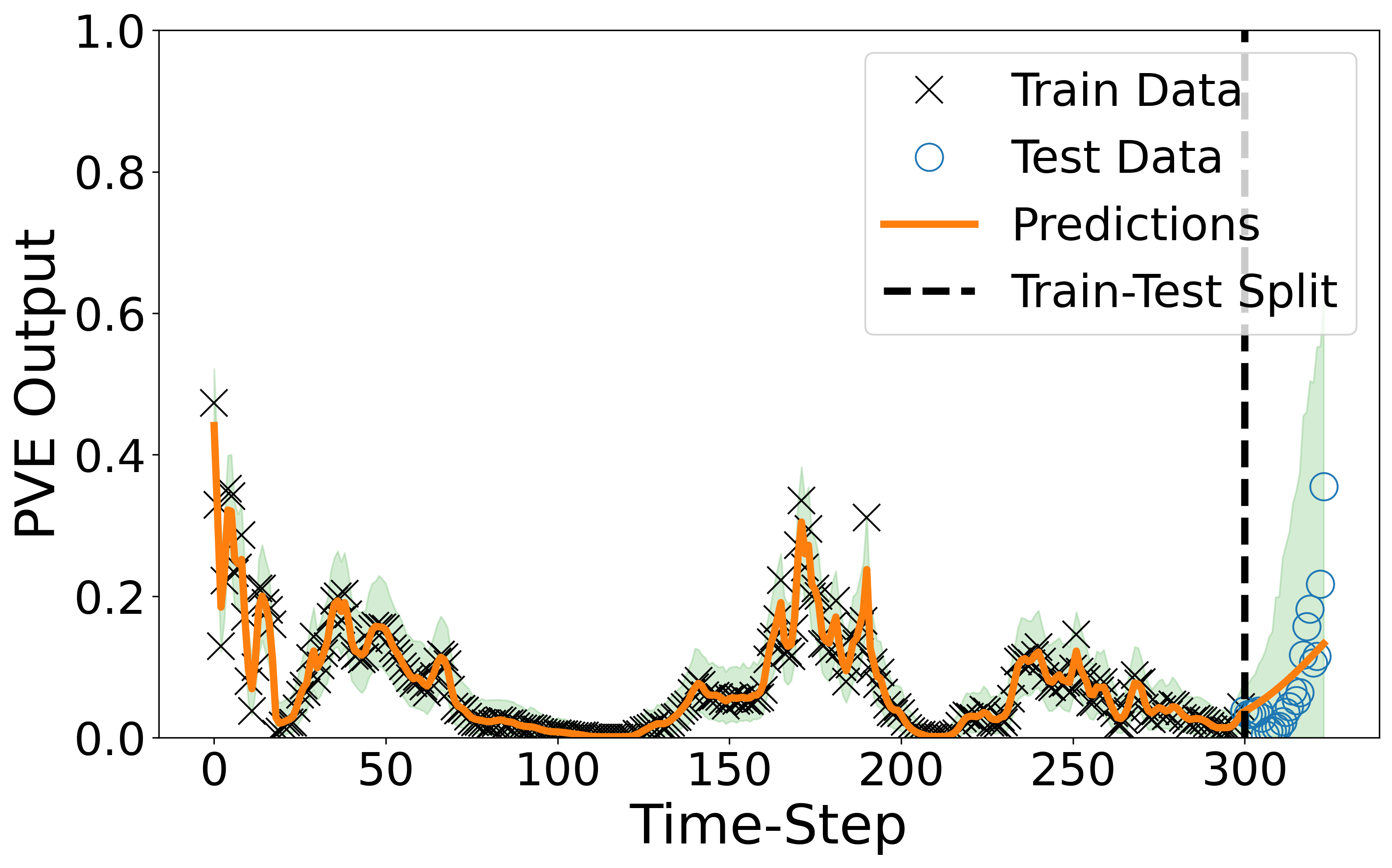}}
    \end{minipage}
    \begin{minipage}{.48\textwidth}
        \centering
        \includegraphics[height=4cm]{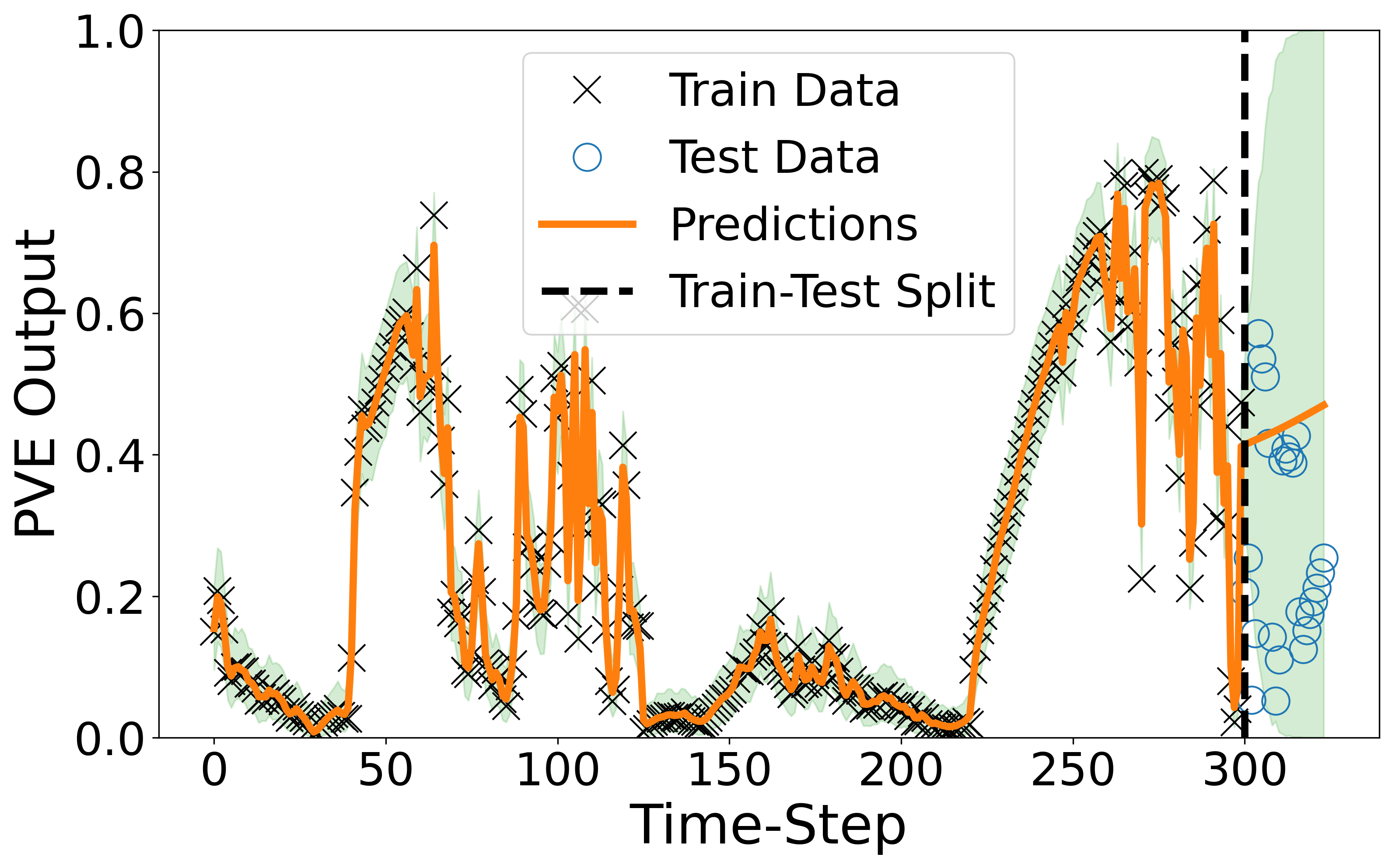}
        \par
        \vspace{15pt}
        \includegraphics[height=4cm]{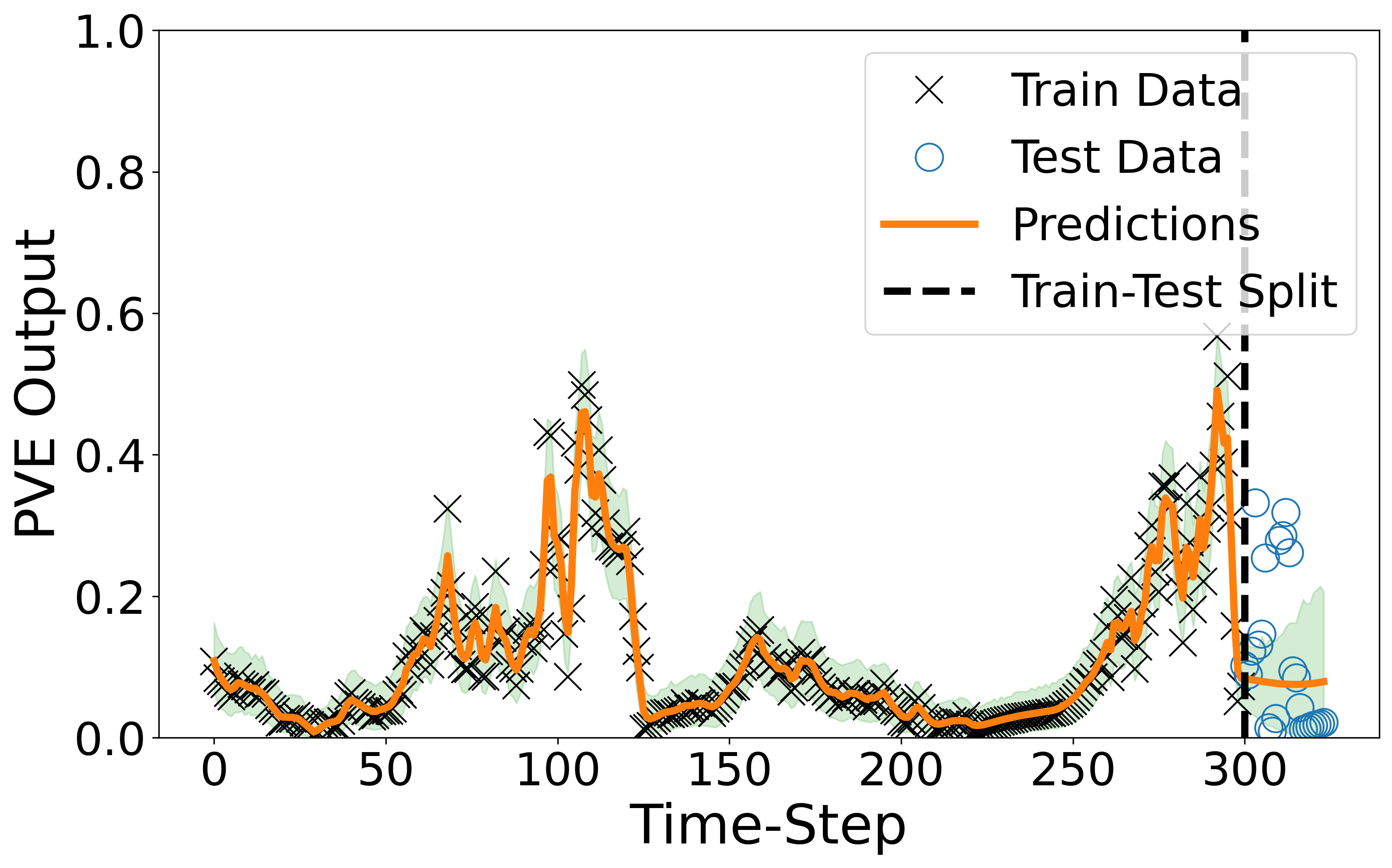}
        \par
        \vspace{15pt}
        \subfloat[Predictions at 2018-02-01]{\includegraphics[height=4cm]{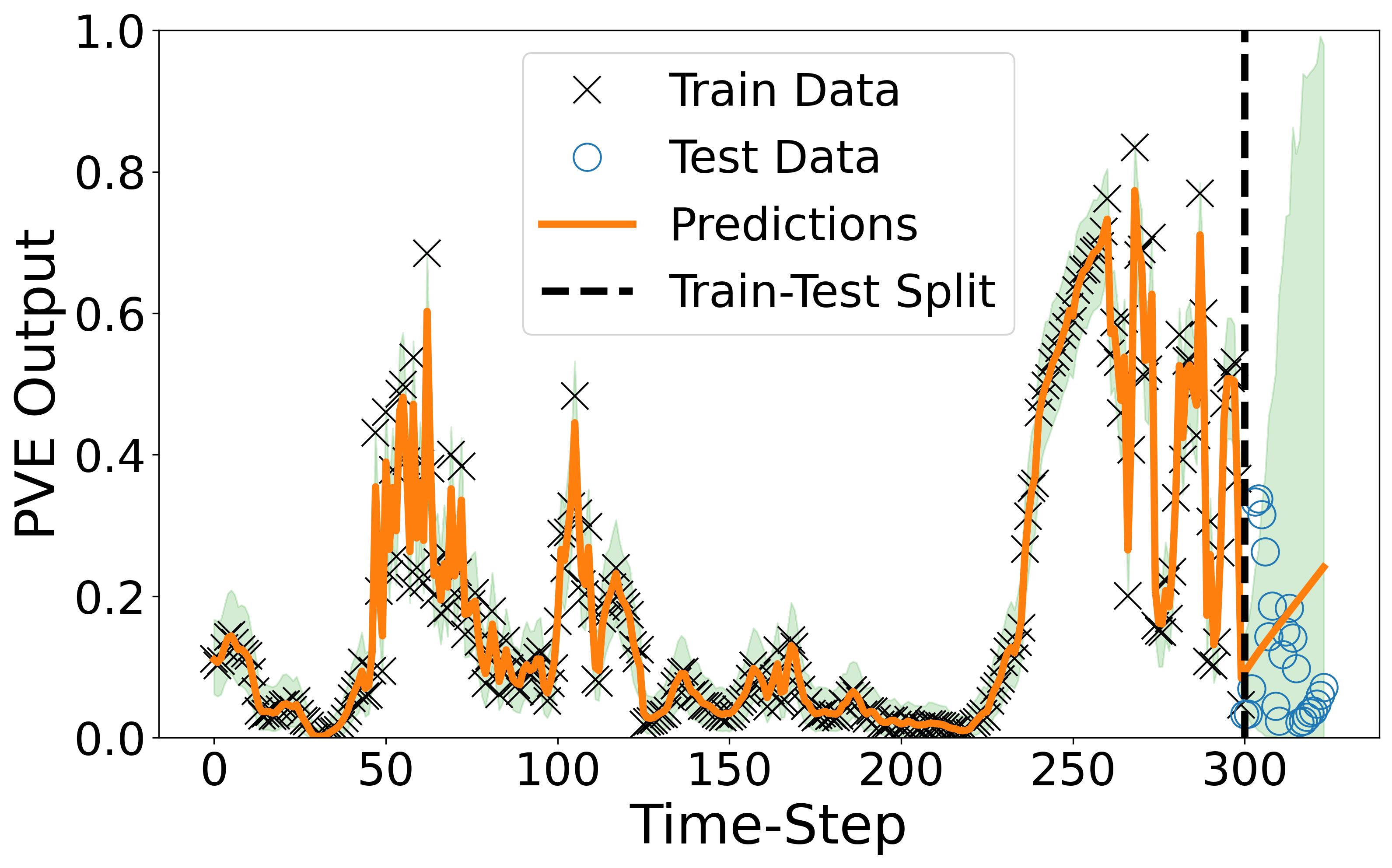}}
        \vspace{15pt}
        \subfloat[Predictions at 2018-02-01]{\includegraphics[height=4cm]{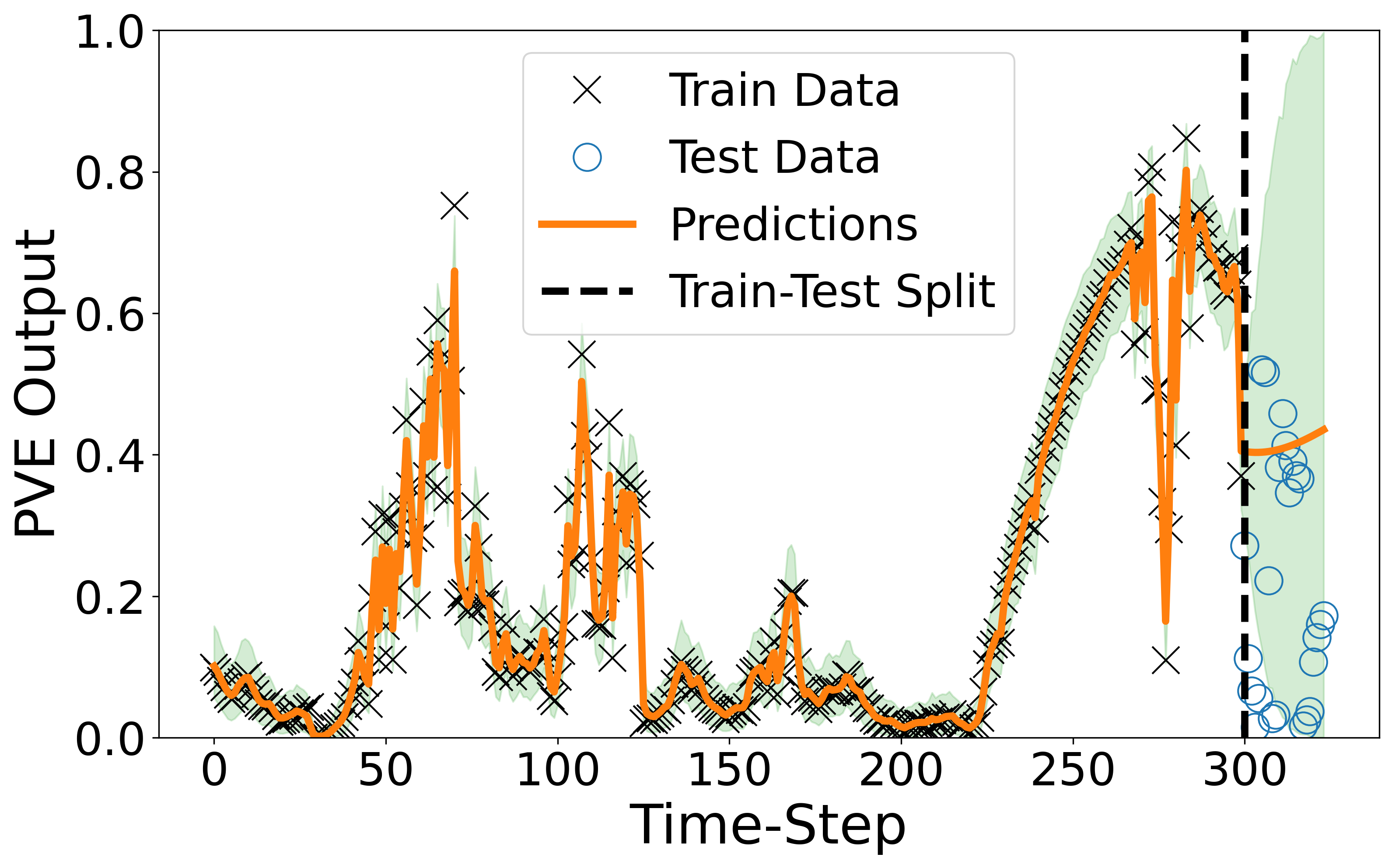}}
    \end{minipage}
    \caption{Predictions from our best performing model at three different PV stations at (a) 2018-01-29, and (b) 2018-02-01. Predictions are generally poor when the timeseries is highly stochastic due to cloud motion and other possible external factors.}
	\label{fig:additional-predictions}
\end{figure*}

In Figure~\ref{fig:additional-predictions}, we plot additional predictions at different PV stations from our best-performing model at two different dates: 2018-01-29 and 2018-02-01. The former was a cloudy day, resulting in overall low PV readings with little-to-moderate variance. The latter was a day with scattered clouds resulting in wildly varying PV readings.

We see that in all stations, the model performs well in the former case. On the other hand, it suffers greatly in the latter case, resulting in high predictive variance, which, while still capturing the ground truth values, has uncertainties too large for it to be useful in downstream tasks.

\end{document}